%% file: main.tex
\PassOptionsToPackage{square,comma,numbers,sort&compress}{natbib}
\documentclass{article}

\usepackage[final]{neurips_2020}
\usepackage[utf8]{inputenc} % allow utf-8 input
\usepackage[T1]{fontenc}    % use 8-bit T1 fonts
\usepackage{hyperref}
\usepackage{url}            % simple URL typesetting
\usepackage{booktabs}       % professional-quality tables
\usepackage{amsfonts}       % blackboard math symbols
\usepackage{nicefrac}       % compact symbols for 1/2, etc.
\usepackage{microtype}      % microtypography
\usepackage{threeparttable}
\usepackage{tablefootnote}
\usepackage{amssymb}  
\usepackage{amsthm}
\usepackage{amsmath}
\usepackage{bm}
\usepackage{mathtools}
\usepackage{caption}
\usepackage{algorithm}
\usepackage{algorithmic}
\usepackage{pifont} 
\usepackage{graphicx}
\usepackage{float}

\usepackage{color}
\usepackage{xcolor}

\input{MACPdef_manu.tex}

\newcommand{\E}{\mathbb{E}}
\newcommand{\KLD}{\text{KL}}

\newcommand{\x}{\bm{x}}
\newcommand{\y}{\bm{y}}

\newcommand{\PPz}{p}
\title{Improving GAN Training with Probability Ratio Clipping and Sample Reweighting}

\author{Yue Wu$^1$,~~ Pan Zhou$^2$,~~ Andrew Gordon Wilson$^3$,~~ Eric P. Xing$^{1,4}$,~~ Zhiting Hu$^{1,5}$ \\
	$^1$Carnegie Mellon University, $^2$Salesforce Research, $^3$New York University\\
	$^4$Petuum Inc., $^5$UC San Diego\\
	{ \texttt \tt wuyueholmes@outlook.com,}\ {\texttt \tt pzhou@salesforce.com,}\ {\texttt\tt aglwilson@gmail.com,}\\ 
	{\texttt \tt epxing@andrew.cmu.edu,}\ {\texttt \tt zhitinghu@gmail.com}
}

\begin{document}

% \makesavenoteenv{tabular}
% \makesavenoteenv{table}
% \theoremstyle{plain}
\newtheorem{claim}{Claim}[section]
\newtheorem{prop}{Proposition}[section]
\newenvironment{psketch}{%
  \renewcommand{\proofname}{Proof sketch}\proof}{\endproof}
\maketitle

\begin{abstract}
% Despite its remarkable success on a wide range of problems related to vision, generative adversarial networks (GANs) can suffer from inferior performance due to unstable training, especially for text generation mainly because of its discrete nature. To solve this issue, we propose a novel variational GAN training framework which enjoys superior  training stability. Inspired by the connection of GANs and reinforcement learning under a variational inference perspective, our framework mainly contains two new techniques:   (1) probability ratio clipping that regularizes generator training to prevent excessively large updates, and (2) a sample re-weighting mechanism that stabilizes discriminator training by downplaying bad-quality fake samples. 
% Moreover, we provide theoretical convergence  analyse of our GAN training framework. 
% Finally, by plugging this variational GAN training framework \pz{into diverse state-of-the-art GAN models ??}, we obtain significant performance improvement  over a range of tasks, including text generation, text style transfer, and image generation. 
Despite success on a wide range of problems related to vision, generative adversarial networks (GANs) often suffer from inferior performance due to unstable training, especially for text generation. To solve this issue, we propose a new variational GAN training framework which enjoys superior  training stability. Our approach is inspired by a connection of GANs and reinforcement learning under a variational perspective. The connection leads to (1) probability ratio clipping that regularizes generator training to prevent excessively large updates, and (2) a sample re-weighting mechanism that improves discriminator training by downplaying bad-quality fake samples.  Moreover, our  variational GAN framework  can provably overcome the training issue in many GANs that an optimal discriminator cannot provide any  informative gradient to training generator. 
%We provide theoretical analysis on the convergence of our approach. 
By plugging the training approach in diverse state-of-the-art GAN architectures, we obtain significantly improved performance over a range of tasks, including text generation, text style transfer, and image generation.\footnote{Code available at: \href{https://github.com/Holmeswww/PPOGAN}{github.com/Holmeswww/PPOGAN}}

\end{abstract}

\section{Introduction}
\label{sec:intro}
Generative adversarial networks (GANs)~\citep{goodfellow2014generative} have achieved remarkable success in image and video synthesis~\cite{DCGAN,brock2018large,mathieu2015deep}. However, it is usually hard to train a GAN well, because the training process is commonly unstable, subject to disturbances and even collapses. 
%caused by imbalance between the strength of the generator and the discriminator. 
To alleviate this issue, substantial efforts have been paid to improve the training stability from different perspectives, 
e.g., divergence minimization \cite{nowozin2016f,nock2017f}, Wasserstein distance with Lipschitz continuity of the discriminator \cite{WGAN,wgangp,wwgan}, energy-based models \cite{zhao2016energy,berthelot2017began}, to name a few. 

In spite of the above progresses, %\cite{nowozin2016f,nock2017f,WGAN,wgangp,wwgan,zhao2016energy,berthelot2017began,che2020your}, 
the instability in training has not been well resolved~\cite{chu2020smoothness}, since it is difficult to well balance the strength of the generator and the discriminator. What is worse, such an instability issue is exacerbated in text generation due to the sequential and discrete nature of text~\cite{fedus2018maskgan,caccia2018language,hu2017toward}. Specifically, the high sensitivity of text generation to noise and the underlying errors caused by sparse discriminator signals in the generated text can often result in destructive updates to both generator and discriminator, enlarging the instability in GANs.  
%\pz{ need to  check.}

% In particular, the high sensitivity of noise in text and the fact that text is generated autoregressively which compounds errors and is updated with sparse discriminator signals, can result in destructive updates to the generator and discriminator, enlarging the instability in GANs. 

% In this work, we develop new techniques to improve the training stability which are broadly applicable to GANs of varied types for image and text generation. The techniques are derived from a  variational   perspective of GANs and the resulting connections to reinforcement learning (in particular, RL-as-inference)~\cite{abdolmaleki2018maximum,levine2018reinforcement,PPO} and other rich literature~\cite{hu2018deep,grover2019bias,burda2015importance}. 

\begin{figure}[t]
%\vskip 0.2in
\begin{center}
\centerline{
\includegraphics[width=0.33\columnwidth]{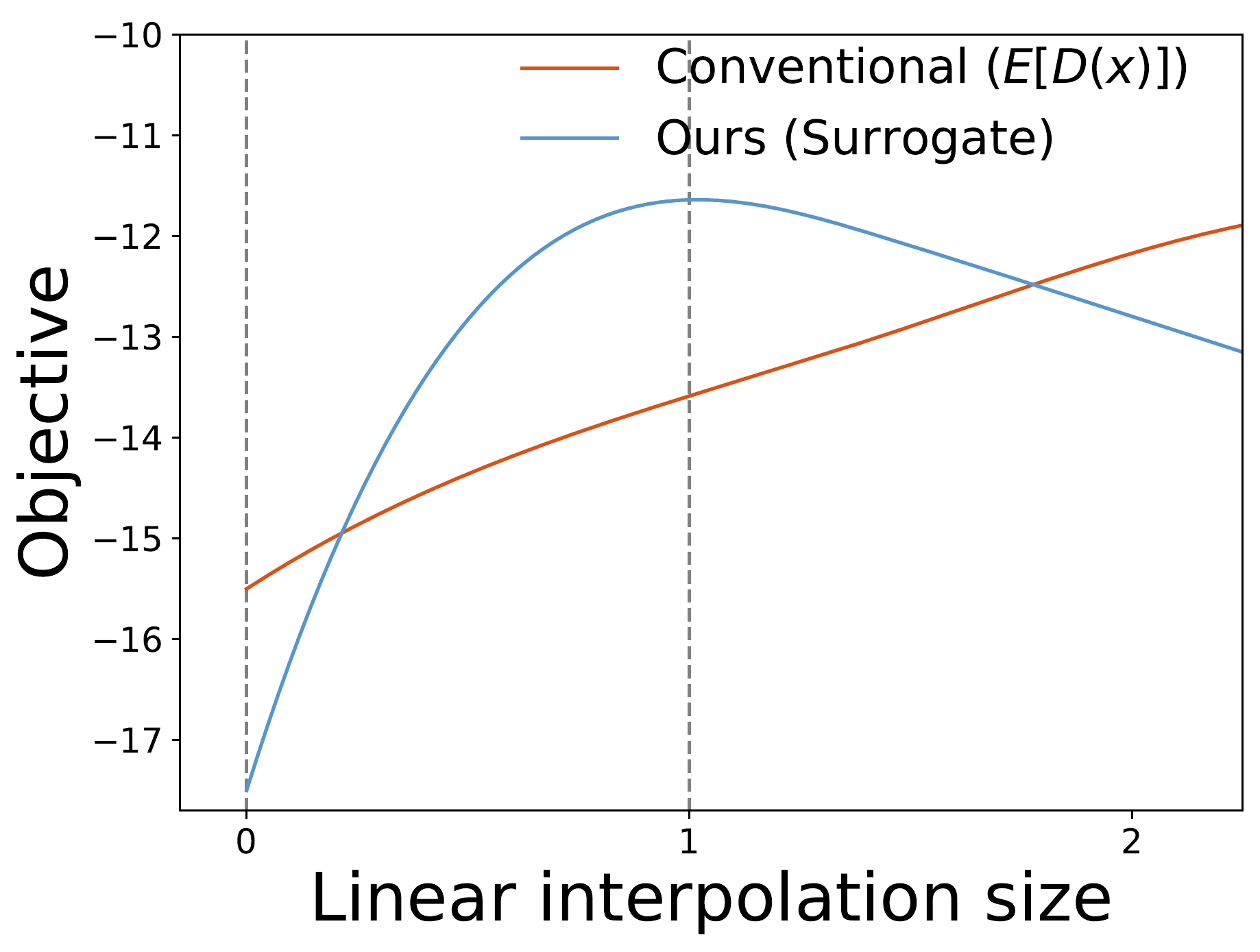}
\includegraphics[width=0.33\columnwidth]{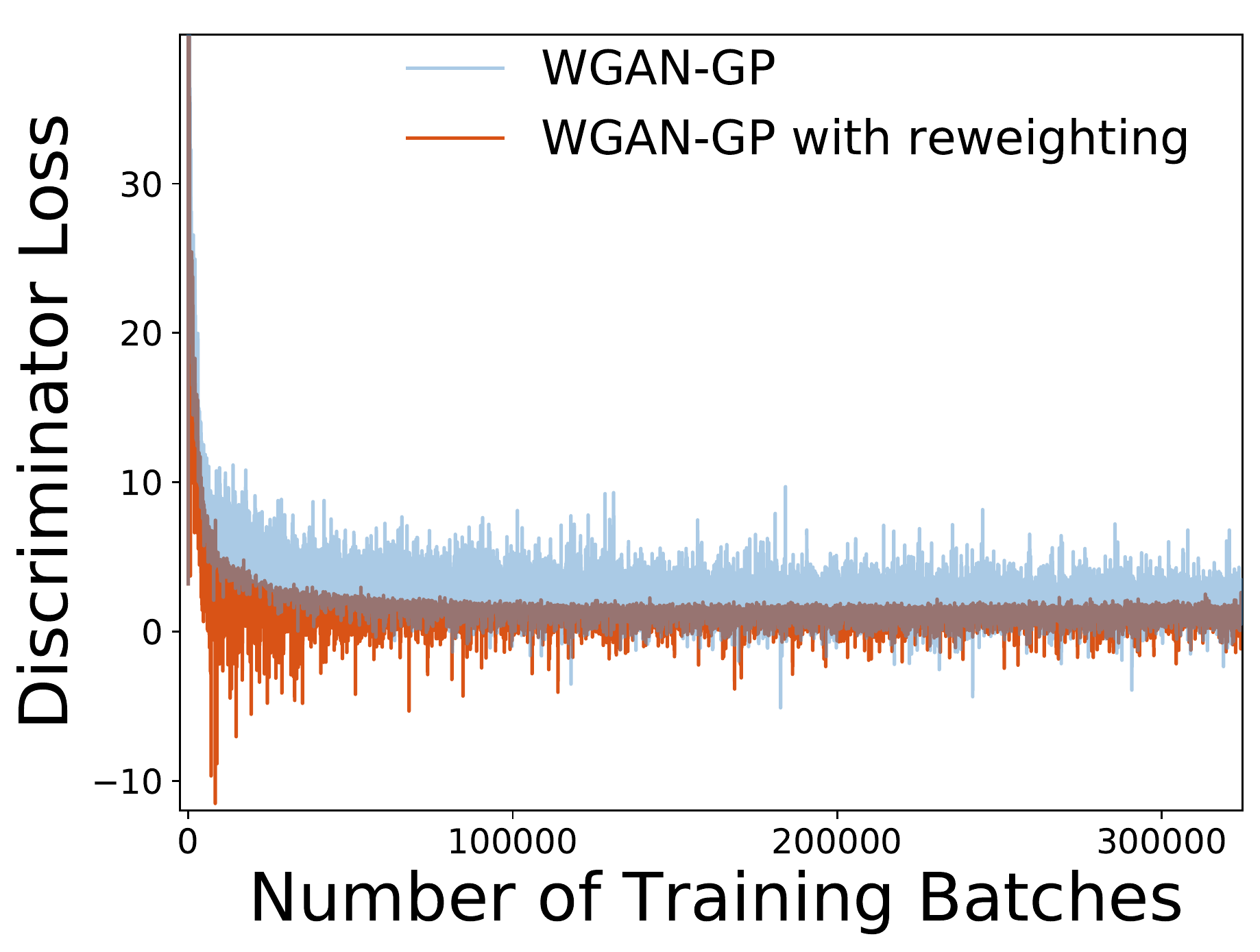}
\includegraphics[width=0.33\columnwidth]{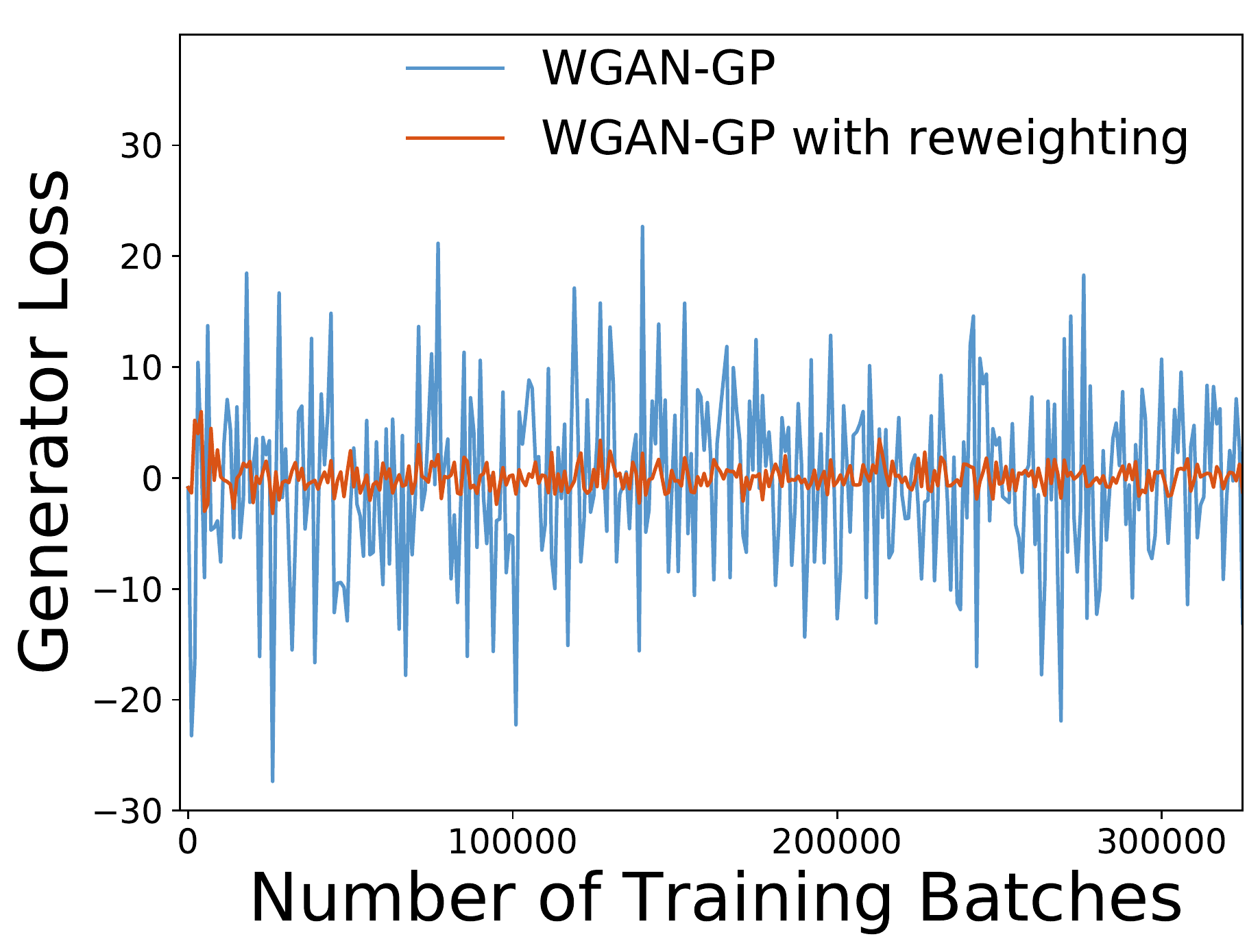}
}
\vspace{-4pt}
\caption{
Illustration of the proposed approach for stabilizing GAN training. Results are from the CIFAR-10 experiment in Sec.\ref{sec:exp:image}.
{\bf Left:} The conventional and surrogate objectives for generator training, as we interpolate between the initial generator parameters $\wm_{old}$ and the updated generator parameters $\wm_{new}$ which we compute after one iteration of training. The $\wm_{new}$ obtains maximal surrogate objective. 
The surrogate objective curve starts decreasing after $x=1$, showing the objective imposes a penalty for having too large of a generator update. In contrast, the conventional objective (for WGAN-GP) keeps increasing with larger generator updates. 
%Note that our surrogate objective penalizes larger step sizes whereas the conventional loss is linear with respect to step size. This plot corresponds to the 10th generator update on the CIFAR10 dataset. 
{\bf Middle and right:} Discriminator and generator losses w/ and w/o sample re-weighting. 
%{\bf Right:} Generator losses w/ and w/o sample re-weighting.
%, where we have subsampled to make the curves clearer. 
WGAN-GP with our re-weighting plugged in shows lower variance in both discriminator and generator losses throughout training. 
%(and achieves better final performance as shown in Sec.\ref{sec:exp:image}). 
}
\label{fig:intro-stability}
\end{center}
% \vskip -0.3in
% \vspace{-10pt}
\vspace{-6mm}
\end{figure}

In this work, we develop a novel variational GAN training framework to improve the training stability, which is broadly applicable to GANs of a variety of architectures for image and text generation. This training framework is derived from a variational perspective of GANs~\citep{hu2018deep} and the resulting connections to reinforcement learning (in particular, RL-as-inference)~\cite{abdolmaleki2018maximum,PPO} and other rich literature~\cite{grover2019bias,hu2017unifying,burda2015importance}. 
Our approach consists of two stabilization techniques, namely, probability ratio clipping and sample re-weighting, for stabilizing the generator and  discriminator respectively.
{\bf (1)} Under the variational perspective, the generator update is subject to a KL penalty on the change of the generator distribution. This KL  penalty closely resembles that in the popular Trust-Region Policy Optimization (TRPO)~\cite{TRPO} and its variant, i.e., Proximal Policy Optimization (PPO)~\cite{PPO}. This connection motivates a simple surrogate objective with a clipped probability ratio between the new generator and the old one. 
The probability ratio clipping discourages excessively large generator updates, and has shown to be effective in the context of stabilizing policy optimization~\cite{PPO}. Figure~\ref{fig:intro-stability} (left) shows the intuition about the surrogate objective, where we can observe the objective value decreases with an overly large generator change and thus imposes regularization on the updates.

% \begin{figure}[ht]
% \vskip 0.2in
% \begin{center}
% \centerline{
% \includegraphics[width=0.33\columnwidth]{figs/reweightD.pdf}
% \includegraphics[width=0.33\columnwidth]{figs/reweightG.pdf}
% \includegraphics[width=0.33\columnwidth]{figs/reweight_norm.pdf}
% }
% \vskip -0.1in
% \caption{Plots comparing the discriminator/generator loss with and without re-weighting with respect to the number of training batches seen. The two trials uses DCGAN architecture without BatchNorm in discriminator and exactly the same set of hyper-parameters on CIFAR 10. Discriminator re-weighted WGAN-GP achieves higher inception score than the original WGAN-GP, while maintaining lower loss variation and consistently lower gradient norm}
% \label{styletransfer_model}
% \end{center}
% % \vskip -0.3in
% \end{figure}

{\bf (2)} When updating the discriminator, the new perspective induces an importance sampling mechanism, which effectively re-weights fake samples by their discriminator scores. Since low-quality samples tend to receive smaller weights, the discriminator trained on the re-weighted samples is more likely to maintain stable performance, and in turn provide informative gradients for subsequent generator updates. Figure~\ref{fig:intro-stability} (middle/right) demonstrates the effect of the re-weighting in reducing the variance of both discriminator and generator losses. 
%This could be particularly helpful in text generation tasks, where the generator tends to produce gibberish samples at the early training stage. 
%Similar importance weighting methods have recently been used in other contexts, such as de-biasing generative models~\cite{grover2019bias} and sampling from energy-based models~\cite{deng2020residual}. Our derivations can be seen as a variant for the new application of improving GANs.

Besides, our variational GAN training framework can provably overcome the training issue~\cite{lipschitz} that an optimal discriminator cannot provide any  informative gradient to training generator. This issue usually occurs in GAN training~\cite{lipschitz}, since the discriminator often converges much faster than the generator. % In this way,  we also provide  theoretical  analysis on the 
%We give theoretical convergence analysis showing the generator receives informative gradient~\cite{lipschitz} from the generator.
%, which guarantees the generalization performance. 
Empirically, we conduct extensive experiments on a wide range of tasks, including text generation, text style transfer, and image generation. Our approach shows significant improvement over state-of-the-art methods, demonstrating its broad applicability and efficacy.

%\section{Background}
%\label{sec:background}

%\vspace{-0.4em}
\section{Related Work}\label{sec:related}
%\vspace{-0.3em}

\textbf{Wasserstein distance, WGAN, and Lipschitz continuity.} 
The GAN framework~\cite{goodfellow2014generative} features two components: a generator $G_\theta$ that synthesizes samples $\x$ given some noise source $\bm{z}$, namely $\x=G_\theta(\bm{z})$ with $\bm{z}\sim p_z(\bm{z})$, and a discriminator that distinguishes generator’s output and real data, which provides gradient feedback to improve the generator's performance. WGAN~\cite{WGAN} improves the training stability of GANs by minimizing the Wasserstein distance $W(p_r, p_\theta)$ between the generation distribution $p_\theta$ (induced from $G_\theta$) and the real data distribution $p_r$.
%
%WGAN~\cite{WGAN} trains a generator network $G(\zm)$, with $\zm\sim \PP_{\zm}$ or some other degenerate probability distributions, by minimizing the Wasserstein distance $W(\PP_r,\PP_{\wm})$ between $\PP_{\wm}$ (the distribution of the generated samples $\{G(\zm)\}$) and the real distribution $\PP_r$ of the observed data points $\{\xm\}$.
%
%The Wasserstein distance is shown better for GAN training than other popular GAN divergences (e.g. Jensen-Shannon divergence) when the target distributions ($\PP_r$, $\PP_{\wm}$) have disjoint support.
%
%Since it is hard to estimate the gradient w.r.t the original formulation of the Wasserstein distance, \citet{WGAN} proposed to optimize the Kantorovich-Rubinstein duality:
Its training loss is formulated as:
\begin{equation}
\min\nolimits_{\bm{\theta}} \max\nolimits_{f \in\mathcal{D}} \EE_{\xm\sim p_r}[f(\x)] - \EE_{\xm\sim p_\theta}[f(\x)],
\label{eq:related:wgan}
\end{equation}
where $\mathcal{D}$ is the set of 1-Lipschitz functions; $f$ acts as the discriminator and is usually implemented by a neural network $f_\phi$.
%Analogous to GANs \cite{goodfellow2014generative}, the $1$-Lipschitz function $f$ acts as the discriminator and guides the learning of generator. This formulation then raises a research topic on enforcing the Lipschitz constraint on the discriminator.
The original resort to enforce the Lipschitz constraint is through weight clipping \cite{WGAN}. WGAN-GP~\cite{wgangp} later improves it by  replacing it with a gradient penalty on the discriminator. CT-GAN~\cite{wwgan} further imposes the Lipschitz continuity constraint on  the manifold of the real data $\xm \sim p_r$. Our approach is orthogonal to these prior works and can serve as a drop-in replacement to stabilize generator and discriminator in various kinds of GANs, such as WGAN-GP and CT-GAN.% as shown in our experiments\pz{ need to be check.}
%we follow WGAN-GP or CT-GAN to enforce Lipschitz continuity and achieves state-of-the-art results on a range of problems.

% However, weight clipping changes the optimization problem and biases the discriminator towards simpler functions \cite{wgangp}. Using the equivalence between Lipschitz continuity and bounded gradient for differentiable functions, \cite{wgangp} resorts to gradient penalty on the discriminator. 
% \[GP|_{\hat{x}}:=\EE_{\hat{x}}\left[(\norm{\nabla_{\hat{x}}f(\hat{x})}_2-1)^2\right]\]
% where $\hat{x}$ is sampled uniformly from the straight line between a real-fake pair sampled from $\PP_r$, $\PP_{\wm}$. However, this loss only encourages bounded gradient at sampled points $\hat{x}$ \cite{wwgan}.

% \cite{wwgan} proposed to further lay the Lipschitz continuity condition over the manifold of the real data $\xm \sim \PP_r$. WGAN-GP \cite{wgangp} and improved WGAN-GP \cite{wwgan} demonstrated stable training and state-of-the-art performance in image generation.

%\paragraph{The effect of Lipschitz continuity}
% Some theory demonstrates  the other benefits of the Wasserstein distance
% There has been some theory around the effect of Lipschitz constraint on GAN discriminators.
Research on the Lipschitz continuity of GAN discriminators have resulted in the theory of ``informative gradients'' \cite{lipschitz,zhou2018understanding}. Under certain mild conditions, a Lipschitz discriminator can provide informative gradient to the generator in a GAN framework: when $p_\theta$ and $p_r$ are disjoint, the gradient $\nabla f^*(\xm)$ of optimal discriminator $f^*$ w.r.t each sample $\xm\sim p_\theta$ points to a sample $\x^* \sim p_r$, which guarantees that the generation distribution $p_{\theta}$ is moving towards $p_r$. We extend the informative gradient theory to our new case and show theoretical guarantees of our approach.
%Under the same assumptions and the discriminator $f$ Lipschitz, \cite{zhou2018understanding} was then able to conclude that when $\PP_{\wm}$ and $\PP_r$ are disjoint, $\nabla f^*(\xm)$ for each sample $\xm\sim\PP_{\wm}$ points to a sample $y\sim\PP_r$, which guarantees that $\PP_{\wm}$ is moving towards $\PP_r$.

%\vspace{-7pt}

\textbf{Reinforcement learning as inference.} 
Casting RL as probabilistic inference has a long history of research~\cite{dayan1997using,deisenroth2013survey,rawlik2013stochastic,levine2018reinforcement,abdolmaleki2018maximum}. 
For example, \citet{abdolmaleki2018maximum} introduced maximum a-posteriori policy optimization from a variational perspective. \citet{tan2018connecting} connected the formulation with other paradigms of learning such as maximum likelihood estimation and data augmentation~\citep{hu2019learning}. 
TRPO~\cite{TRPO} is closely related to this line by using a KL divergence regularizer to stabilize standard RL objectives. PPO~\cite{PPO} further proposed a practical clipped surrogate objective that emulates the regularization. Our approach draws on the connections to the research, particularly the variational perspective and PPO, to improve GAN training.
%\citet{peters2010relative} introduced a relative entropy regularization to reduce information loss during policy updates. 
%Of particular relevance to our work is 

%\vspace{-7pt}

%\textbf{Algorithm re-purposing.} 

\textbf{Other related work.} 
Importance re-weighting has been adopted in different problems, such as learning knowledge constraints~\cite{hu2018deep}, and improving VAEs~\cite{burda2015importance} and GANs~\citep{hu2017unifying,song2019bridging}. 
%proposed a KL-WGAN objective based on variational divergence minimization and demonstrated the experimental benefits of a re-weighted objective. 
We derive from the variational perspective which leads to re-weighting and clipping in the new context of GAN training stabilization. Our approach is orthogonal to and can be combined with other stabilization techniques such as large-batch training \citep{brock2018large} and parameter averaging (EMA) \citep{yaz2018unusual,brock2018large}. 
%\citet{che2020your} 
%GANs have also been studied from other perspectives. 
%Of relevance to our work is 

% \subsection{TRPO and PPO}
% % \pz{I think it is not necessary to introduce TRPO and PPO here, since 1) here you focus on more GAN and 2) in the method part, you only use their optimization trick (you only need to mention them in the method part).}
% Vanilla policy gradient methods in reinforcement learning has poor data efficiency and robustness. In trust region policy optimization (TRPO) \cite{TRPO}, an objective function (the “surrogate” objective) is maximized subject to a constraint on the size of the policy update measured by KL divergence. The theory behind TRPO guarantees monotonic improvement of performance, and high data efficiency. However, such trust region methods are relatively complicated and incompatible with architectures that includes noise (such as dropout) or parameter sharing. 

% PPO \cite{PPO} proposed a clipped surrogate objective to mimic the behavior of a KL constrained loss.
% \begin{equation}\label{TRPO}
% \begin{split}
%   & L^{CLIP}(\wm) \\
% = & \hat{\EE}_t\left[\min\left(r_t(\theta) \hat{A_t}, clip\left(r_t(\theta), 1-\epsilon, 1+\epsilon\right) \hat{A_t}\right)\right]
% \end{split}
% \end{equation}
% In practice, PPO demonstrates good scalability, with performance comparable with TRPO in most cases. PPO also improves sample efficiency and sometimes allows an agent trajectory to be reused several times even in an online learning scenario.

%\fi

\section{Improving GAN Training}
\subsection{Motivations}
Our approach is motivated by connecting GAN training with the well-established RL-as-inference methods~\cite{abdolmaleki2018maximum,levine2018reinforcement,tan2018connecting} under a variational perspective. 
The connections enable us to augment GAN training with existing powerful probabilistic inference tools as well as draw inspirations from the rich RL literature for stable training. In particular, the connection to the popular TRPO~\cite{TRPO} and PPO~\cite{PPO} yields the probability ratio clipping in generator training that avoids destructive updates  (Sec.\ref{sec:method:gen}), and the application of importance sampling estimation gives rise to sample re-weighting for adaptive discriminator updates (Sec.\ref{sec:method:dis}). The full training procedure is summarized in Alg.\ref{alg:opt}.

Specifically, as described in Sec.\ref{sec:related}, the conventional WGAN formulation for updating the generator $p_\theta(\x)$ maximizes the expected discriminator score $\E_{p_\theta}[ f_\phi(\x) ]$, where $f_\phi$ is the Lipschitz-continuous discriminator parameterized with $\bm{\phi}$. The objective straightforwardly relates to policy optimization in RL by seeing $p_\theta$ as a policy and $f_\phi$ as a reward function. Thus, inspired by the probabilistic inference formulations of policy optimization~\cite{abdolmaleki2018maximum,hu2018deep,tan2018connecting}, here we transform the conventional objective by introducing a non-parameterized auxiliary distribution $q(\x)$ and defining a new variational objective:
\begin{equation}
\small
\begin{split}
    \mathcal{L}(\bm{\theta}, q) = \E_q[ f_\phi(\x) ] - \KLD\left( q(\x) \| p_\theta(\x) \right),
\end{split}
\label{eq:var-obj}
\end{equation}
where $\KLD$ is the KL divergence.
Intuitively, we are maximizing the expected discriminator score of the auxiliary $q$ (instead of generator $p_\theta$), and meanwhile encouraging the generator to stay close to $q$. We note that \citet{hu2018deep} have also related the above objective to GANs, with the different goal of integrating structured knowledge with deep generative modeling. 

As we shall see in more details shortly, the
%new treatment of the generator updates 
new formulation allows us to take advantage of off-the-shelf inference methods, which naturally leads to new components to improve the GAN training. Maximization of the above objective is solved by the expectation maximization (EM) algorithm~\cite{neal1998view} which alternatingly optimizes $q$ at E-step and optimizes $\bm{\theta}$ at M-step. More specifically, at each iteration $t$, given the current status of generator parameters $\bm{\theta}=\bm{\theta}^{(t)}$, the E-step that maximizes $\mathcal{L}(\bm{\theta}^{(t)}, q)$ w.r.t $q$ has a closed-form solution:
\begin{equation}
\small
\begin{split}
q^{(t)}(\x) = \frac{p_{\theta^{(t)}}(\x)\exp\{ f_\phi(\x) \}}{Z_\phi},
\end{split}
\label{eq:q-sol}
\end{equation}
where $Z_\phi = \int_x p_{\theta^{(t)}}(\x)\exp\{ f_\phi(\x) \}$ is a normalization term that  depends on the discriminator parameters $\bm{\phi}$.
We elaborate on the M-step in the following subsections, where we continue to develop the practical procedures for updating the generator and the discriminator, respectively.
%Building upon the above derivations, in the following section, we continue to develop the procedures for updating the generator and the discriminator, respectively. In the following section, we will elaborate on the M-step. 

\subsection{Generator Training with Probability Ratio Clipping}\label{sec:method:gen}

The M-step optimizes $\mathcal{L}(\bm{\theta}, q^{(t)})$ w.r.t $\bm{\theta}$, which is equivalent to minimizing the KL divergence term in Eq.\eqref{eq:var-obj}. However, since the generator $p_\theta$ in GANs is often an \emph{implicit} distribution that does not permit evaluating likelihood, the above KL term (which involves evaluating the likelihood of samples from $q$) is not applicable. We adopt an approximation, which has also been used in the classical wake-sleep algorithm~\cite{hinton1995wake} and recent work~\cite{hu2018deep}, by minimizing the \emph{reverse} KL divergence as below. With Eq.\eqref{eq:q-sol} plugged in, we have:
%Plugging this reverse KL divergence into Eq.\eqref{eq:q-sol}, we have:
%\yw{@zhiting, please add one line to point to pf in appendix}
\begin{equation}
\small
\begin{split}
\min\nolimits_\theta \KLD\left( p_\theta(\x) \| q^{(t)}(\x) \right) = \min\nolimits_\theta - \E_{p_\theta}\left[ f_\phi(\x) \right] + \KLD\left( p_\theta(\x) \| p_{\theta^{(t)}}(\x) \right).
\end{split}
\label{eq:g-loss}
\end{equation}
As proven in the appendix, this reverse KL approximation does not change the optimization problem in Eq.\eqref{eq:var-obj}. 
The first term on the right-hand side of Eq.\eqref{eq:g-loss} recovers the conventional objective of updating the generator. Of particular interest is the second term, which is a new KL regularizer between the generator $p_\theta$ and its ``old'' state $p_{\theta^{(t)}}$ from the previous iteration. The regularizer discourages the generator from changing too much between updates, which is useful to stabilize the stochastic optimization procedure. The regularization closely resembles to that of TRPO/PPO, where a similar KL regularizer is imposed to prevent uncontrolled policy updates and make policy gradient robust to noises. Sec.\ref{sec:theory} gives  analysis on the KL-regularized generator updates.

In practice, directly optimizing with the KL regularizer can be infeasible due to the same difficulty with the implicit distribution as above. Fortunately, PPO~\cite{PPO} has presented a simplified solution that emulates the regularized updates using a clipped surrogate objective, which is widely-used in RL. We import the solution to our context, leading to the following practical procedure of generator updates.

\textbf{Probability Ratio Clipping.}  
Let $r_t$ denote the probability ratio $r_t(\bm{\theta})=\frac{p_\theta(\x)}{p_{\theta^{(t)}}(\x)}$ which measures the difference between the new and old generator distributions.
For instance, $r_t(\bm{\theta}^{(t)}) = 1$. 
The clipped surrogate objective for updating the generator, as adapted from PPO, is:
\begin{equation}
\small
\begin{split}
\mathcal{L}^{CLIP}(\bm{\theta}) = \E_{p_\theta}\left[ \min\left( r_t(\bm{\theta}) f_\phi(\x),~ r_t^{clip}(\bm{\theta}) f_\phi(\x) \right) \right],
\end{split}
\label{eq:obj-clip}
\end{equation}
where $r_t^{clip}(\bm{\theta}) = \text{clip}\left(r_t(\bm{\theta}), 1-\epsilon, 1+\epsilon\right)$ clips the probability ratio, so that moving $r_t(\bm{\theta})$ outside of the interval $[1-\epsilon, 1+\epsilon]$ is discouraged. Taking the minimum puts a ceiling on the increase of the objective. Thus the generator does not benefit by going far away from the old generator.

Finally, to estimate the probability ratio $r_t(\bm{\theta})$ when $p_\theta$ is implicit, we use an efficient approximation similar to \cite{MLGAN,grover2019bias} by introducing a binary classifier $C$ trained to distinguish real and generated samples. Assuming an optimal classifier $C$ which has $p_\theta(\x) = \frac{1-C(\x)}{C(\x)}p_r(\x)$~\cite{goodfellow2014generative,MLGAN}, we approximate $r_t$ by: 
%keeping a historical version ($C$) of $C^{(t)}$ at time-step $t$:
\begin{equation}
\small
\begin{split}
r_t(\bm{\theta}) = \frac{p_\theta(\x)}{p_{\theta^{(t)}}(\x)} \approx \frac{(1-C(\x))\cdot C^{(t)}(\x)}{(1-C^{(t)}(\x))\cdot C(\x)},
\end{split}
\label{eq:r-approx}
\end{equation}
where $C^{(t)}(\x)$ denotes the classifier at the $t$-th iteration. 
%Note that after plugging the rightmost expression into Eq.\eqref{eq:obj-clip}, gradient can propagate through $C$ to $\bm{\theta}$ since $\x=G_\theta(\bm{z})$. 
Note that the rightmost expression depends on $\bm{\theta}$ because $\x$ is the output of the generator, i.e., $\x=G_\theta(\bm{z})$.
In practice, during the phase of generator training, we maintain $C$ by fine-tuning it for only one iteration every time after $\bm{\theta}$ is updated (Alg.\ref{alg:opt}). Thus the maintenance of $C$ is cheap. We give more details of the configuration of $C$ in the appendix. In the cases where an explicit generative model is used (e.g., a language model for text generation), the probability ratio $r_t$ can directly be evaluated by definition without the need of $C$, though in our text generation experiments (Sec.\ref{sec:exp:text}) we still used $C$ for approximating $r_t$.
%\hzt{More details in appendix.} 

% \begin{equation}
% \begin{split}
% r_t(\theta) & = \frac{P_G(x)}{P_{G_0}(x)}\\
%             & = \frac{\frac{1-D'(x)}{D'(x)} p_r (x)}{\frac{1-D_0'(x)}{D_0'(x)}p_r(x)}\\
%             & = \frac{\frac{1-D'(x)}{D'(x)}}{\frac{1-D_0'(x)}{D_0'(x)}}
% \end{split}
% \label{eq:obj-clip}
% \end{equation}

\subsection{Discriminator Training with Sample Re-weighting}\label{sec:method:dis}

We next discuss the training of the discriminator $f_\phi$, where we augment the conventional training with an importance weighting mechanism for adaptive updates. 
Concretely, given the form of the auxiliary distribution solution $q^{(t)}$ in Eq.\eqref{eq:q-sol}, we first draw from the recent energy-based modeling work~\cite{kim2016deep,hu2018deep} and propose to optimize $\bm{\phi}$ by maximizing the data log-likelihood of $q^{(t)}$, $\mathcal{L}(\bm{\phi})=\E_{p_r}[\log q^{(t)}(\x)]$. By taking the gradient, we have: %\hzt{Lipschitz}
\begin{equation}
\small
\begin{split}
    \nabla_\phi \mathcal{L}(\bm{\phi}) = \nabla_\phi \Big( \E_{p_r}\left[ f_\phi(\x)  \right] - \log Z_\phi \Big) 
    = \E_{p_r}\left[ \nabla_\phi  f_\phi(\x)  \right] -  \E_{q^{(t)}}\left[ \nabla_\phi f_\phi(\x) \right].
\end{split}
\label{eq:d-grad}
\end{equation}
%To encourage training convergence, we set $f_\phi$ to be from the class of $1$-Lipschitz functions. \yw{edited}
%After this updating, we  clip parameter $\phi$ to ensure that $f_\phi$ has upper bounded Lipschitz constant \pz{need check}. 
We can observe that the resulting form resembles the conventional one (Eq.\ref{eq:related:wgan}) as we are essentially maximizing $f_\phi$ on real data while minimizing $f_\phi$ on fake samples. An important difference is that here fake samples are drawn from the auxiliary distribution $q^{(t)}$ instead of the generator $p_\theta$. This  difference leads to the new sample re-weighting component as below. Note that, as in WGAN (Sec.\ref{sec:related}), we maintain $f_\phi$ to be from the class of $1$-Lipschitz functions, which is necessary for the convergence analysis in Sec.\ref{sec:theory}. In practice, we can use gradient penalty \cite{wgangp,wwgan} for the Lipschitz continuity.

\textbf{Sample Re-weighting.}  
We use the tool of importance sampling to estimate the expectation under $q^{(t)}$ in Eq.\eqref{eq:d-grad}. Given the multiplicative form of $q^{(t)}$ in Eq.\eqref{eq:q-sol}, similar to~\cite{abdolmaleki2018maximum,hu2018deep,deng2020residual}, we use the generator $p_{\theta^{(t)}}$ as the proposal distribution. This leads to 
\begin{equation}
\small
\begin{split}
\E_{q^{(t)}}\left[ \nabla_\phi f_\phi(\x) \right] = \E_{p_{\theta^{(t)}}}[ \exp\{f_\phi(\x)\} \cdot \nabla_\phi f_\phi(\x) ] \ /\  Z_\phi.
\end{split}
\label{eq:d-weight}
\end{equation}
Note that $Z_\phi$ is the normalization factor defined in Eq.\eqref{eq:q-sol}. Thus, fake samples from the generator are weighted by the exponentiated discriminator score when used to update the discriminator. Intuitively, the mechanism assigns higher weights to samples that can fool the discriminator better, while low-quality samples are downplayed to avoid destructing the discriminator performance. It is worth mentioning that similar importance weighting scheme has been used in~\cite{hu2017unifying,MLGAN} for generator training in GANs, and~\cite{burda2015importance} for improving variational auto-encoders. Our work instead results in a re-weighting scheme in the new context of discriminator training.

%\subsection{The Algorithm}\label{sec:alg}
The algorithm below summarizes the proposed training procedure for the generator and discriminator.

\begin{algorithm}[!h]
%\small
\centering
\caption{\small GAN Training with Probability Ratio Clipping and Sampling Re-weighting}
\label{alg:opt}
\begin{algorithmic}[1]
%\REQUIRE The generator $p_\theta(\bm{x})$ \\
%\quad\ \  The (set of) constraints $f_\phi(\bm{x})$  \\
\STATE Initialize the generator $p_\theta$, the discriminator $f_\phi$, and the auxiliary binary classifier $C$
\FOR {$t \gets 1$ to $T$}
    \FOR {certain number of steps}
	    \STATE Update the discriminator $f_\phi$ with sample re-weighting through Eqs.\eqref{eq:d-grad}-\eqref{eq:d-weight}, and maintain $f_\phi$ to have upper-bounded Lipschitz constant through, e.g., gradient penalty~\cite{wgangp}.
	\ENDFOR
	\FOR {certain number of steps}
	    \STATE Finetune the real/fake binary classifier $C$ (for 1 step)
	    \STATE Estimate probability ratio $r_t(\bm{\theta})$ using $C$ through Eq.\eqref{eq:r-approx}
	    \STATE Update the generator $p_\theta$ with probability ratio clipping through Eq.\eqref{eq:obj-clip}
    \ENDFOR
\ENDFOR
%\ENSURE The trained generator $p_{\theta^*}$
\end{algorithmic}
\end{algorithm}

%\vspace{-0.8em}
\subsection{Theoretical Analysis}\label{sec:theory}
%\vspace{-0.3em}
% In this section, we show that the generator distribution $\PPz_\theta$ with our training approach can converge to the real data distribution $\PPz_r$. 
To provide theoretical insight on the performance of our method, we prove that our framework holds the same guarantees as WGAN-GP~\citep{wgangp} and Lipschitz GANs~\citep{lipschitz}. Formally, we show that the method is fully compatible with \emph{Proposition 1} in \citep{wgangp} and \emph{Theorem 2} in \citep{lipschitz}, which provides rigorous analysis on GANs with Lipschitz discriminators and concludes 1) informative gradient pushes the generator distribution to the real data distribution and 2) the only Nash-equilibrium is $p_{\theta}=p_{r}$. Note that the theorems do not guarantee distributional convergence of $p_{\theta}$ to $p_r$, same as in \citep{wgangp,lipschitz}.

Our analysis is based on the reverse KL updates for the generator (Eq.\ref{eq:g-loss}), while the probability ratio clipping serves as a practical emulation for the updates. We begin by adapting \emph{Proposition 1} in \citet{wgangp} to our problem: 
\begin{prop} Let $\PPz_r$ and $q$ be two distributions in $X$, a compact metric space. Then, there is a $1$-Lipschitz function $f^*$ which is the optimal solution of $$\max_{\norm{f}_L \leq 1} \EE_{\xm \sim \PPz_r} \left[f(\xm)\right] - \EE_{\xm\sim q} \left[f(\xm)\right]$$
Let $\pi^*$ be the optimal coupling between $\PPz_r$ and $q$, defined as the minimizer of: $W(\PPz_r, q) = \inf_{\pi \in \Pi(\PPz_r,q)} \EE_{(\x,\y)\sim \pi} \left[\norm{\x - \y}\right]$ where $\Pi(\PPz_r, q)$ is the set of joint distributions $\pi(\x, \y)$ whose marginals are $\PPz_r$ and $q$, respectively. Then, if $f^*$ is differentiable, $\pi^*(\x = \y) = 0$, and $\x_\tau = \tau \x + (1-\tau)\y$ with $0 \leq \tau \leq 1$, it holds that 
$\PP_{(\x,\y)\sim\pi^*}\left[\nabla f^*(\x_\tau)=\frac{\y-\x_\tau}{\norm{\y-\x_\tau}}\right] = 1$. \label{wgangpprop}
\end{prop}
Proposition~\ref{wgangpprop} indicates that in presence of an optimal discriminator $f^*$, given any sample $\ym$ drawn from the variational distribution $q$, there exists a sample $\xm$ drawn from real data distribution $\PPz_r$ such that $\nabla_{\xm} f^*(\xm_\tau)\!=\!\frac{\ym-\xm_\tau}{\norm{\ym-\xm_t}}$ for all linear interpolations  $\xm_\tau  =  \tau \xm + (1-\tau)\ym$ with $0 \leq \tau \leq 1$. Therefore, an optimal discriminator $f^*$ can provide informative gradient to update $q$ and push $q$ towards to the real distribution $\PPz_r$. %On the other hand, by minimizing KL distance between the generator distribution $p$ and the auxiliary distribution $q$, the distribution $p$ will become close to $q$ and thus the real distribution $\PPz_r$, which means that the generator is still trainable even  with  an optimal discriminator $f^*$ .
%%%OLD
%The proposition shows that the optimal discriminator $f^*$ provides informative gradient \cite{zhou2018understanding} from the auxiliary distribution $q$ towards the real data distribution $\PPz_r$. We then generalize the conclusion to $\PPz_{\theta}$ by considering correlation between $q$ and $\PPz_{\theta}$.
%

By the definition of $q$ with respect to $\PPz_{\theta}$ in Eq.\eqref{eq:q-sol}, the support of $\PPz_{\theta}$ and $q$ are the same; namely, given any $\x\sim \PPz_{\theta}, \y\sim \PPz_r$, we also have $q(\x) \neq 0$. Therefore, for all $\xm \sim \PPz_{\theta}$, $\xm$ is also a valid sample from $q$, the $f^*$ in {Proposition \ref{wgangpprop}} provides informative gradient with respect to $\x_\tau = \tau \x + (1-\tau)\y, \forall \tau\in [0,1]$:
$\PP_{(\x,\y)\sim\pi^*}\left[\nabla f^*(\x_\tau)=\frac{\y-\x_\tau}{\norm{\y-\x_\tau}}\right] = 1$
Therefore, assuming $f^*$ is the optimal discriminator to \eqref{eq:d-grad}, optimizing Eq.\eqref{eq:g-loss} can provide informative gradient that points the generator $p_\theta$ toward $p_r$.

\section{Experiments} \label{sec:results}
%\vspace{-0.3em}
%
We conduct extensive experiments on three unsupervised generation tasks, including image generation, text generation, and text style transfer. The three tasks apply GANs to model different data modalities, namely, image, text, and neural hidden representations, respectively. Our approach consistently offers improvement over the state-of-the-arts on all tasks. See appendix for all experimental details.

% To benchmark our framework, we apply our loss to competitive benchmarks: text style transfer and image generation. Moreover, we also We also conduct ablation studies on the effects of the two separate parts of our model: (1) discriminator weighted samples (2) clipped surrogate loss.

\begin{figure}[t]
\begin{minipage}{0.47\textwidth}
% \begin{table}[H]
%  \vskip - 0.23in
\vspace{-0.8em}
\setlength{\tabcolsep}{2.2pt}  
\renewcommand{\arraystretch}{0.85} 
\footnotesize
  \centering
  \begin{tabular}{@{}lll@{}}
    \toprule
    {\bf Method}     & {\bf IS} ($\uparrow$) & {\bf FID} ($\downarrow$) \\
    \cmidrule{1-3}
    Real data & 11.24$\pm$.12 & 7.8\\
    % \cmidrule(r){1}
    \cmidrule{1-3}
    % \textbf{-Standard CNN-} \\
    % Weight clipping (\citeyear{WGAN}) & 6.41$\pm$.11 & 42.6 \\
    % GAN-GP (\citeyear{wgangp}) & 6.93$\pm$.08 & 37.7 \\
    % WGAN-GP (\citeyear{wgangp}) & 6.68$\pm$.06 & 40.2 \\
    % Batch Norm. (\citeyear{SNGAN}) & 6.27$\pm$.10 & 56.3 \\
    % Layer Norm. (\citeyear{SNGAN}) & 7.19$\pm$.12 & 33.9 \\
    % Weight Norm. (\citeyear{SNGAN}) & 6.84$\pm$.07 & 34.7 \\
    % Orthonormal (\citeyear{SNGAN}) & 7.40$\pm$.12 & 29.0 \\
    % SN-GANs (\citeyear{SNGAN}) & 7.58$\pm$.12 & 25.5 \\
    % \midrule
    % \textbf{-ResNet-}\\
    %Orthonormal (\citeyear{SNGAN}) & 7.92$\pm$.04 & 23.8$\pm$.58 \\
    %DCGAN (\citeyear{DCGAN}) & 6.64$\pm$.14 & - \\
    %LR-GANs (\citeyear{LRGAN}) & 7.17$\pm$.07 & - \\
    %Denoising (\citeyear{denoise}) & 7.72$\pm$.13 & - \\
    WGAN-GP (\citeyear{wgangp}) & 7.86$\pm$.08 & - \\
    CT-GAN (\citeyear{wwgan}) &  8.12$\pm$.12 & - \\
    SN-GANs (\citeyear{SNGAN}) & 8.22$\pm$.05 & 21.7$\pm$.21 \\
    WGAN-ALP (\citeyear{wganalp}) & 8.34$\pm$.06 & 12.96$\pm$.35\\
    SRNGAN (\citeyear{srngan}) & 8.53 $\pm$.04 & 19.83\\
    %AutoGAN (\citeyear{gong2019autogan}) & 8.55$\pm$.10 & 12.42\\\cmidrule{1-3}
    Ours {(re-weighting only)} & 8.45$\pm$.14 & 13.21$\pm$.60 \\
    Ours (full) & \textbf{8.69$\pm$.13} &\textbf{10.70$\pm$.10}\\
    \bottomrule
  \end{tabular}
  \vspace{-5pt}
  \captionof{table}{CIFAR-10 results. 
  %We copied the results of previous GAN models from the corresponding papers.
  %We report the average and standard deviation over 3 runs.
  Our method is run 3 times for  average and  standard deviation. 
  %and  are averaged over 3 runs $\pm$ one standard deviation.
  } %TODO
  \label{tab:exp:image:IS_FID}
\vskip -0.4in
% \end{table}
\end{minipage}
\hfill
\begin{minipage}{0.51\textwidth}
% \begin{figure}[t]
% \vskip 0.1in
\begin{center}
\centerline{
\includegraphics[width=0.9\columnwidth]{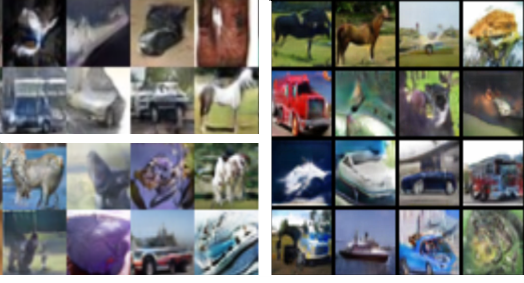}
}
\vskip -0.1in
\captionof{figure}{Generated samples by WGAN-GP (top-left), CT-GAN (bottom-left), and ours (right). 
%Our generated images (right) have higher quality than the samples generated by unsupervised WGAN-GP (ResNet) \cite{wgangp} (top-left) and CT-GAN (ResNet) \cite{wwgan} (bottom-left).
% unsupervised WGAN-GP (ResNet) \cite{wgangp} (top-left) and CT-GAN (ResNet) \cite{wwgan} (bottom-left) compared to our generated (right) samples. Our generated samples are of better visual quality than the samples generated by the baselines.
}
\label{cifar10_combined}
%\label{samples}
\end{center}
%\vskip -0.4in
% \end{figure}
\end{minipage}
\vspace{-15pt}
\end{figure}

\begin{figure}[t]
% \vskip 0.1in
\begin{center}
%\centerline{
\includegraphics[width=0.40\columnwidth]{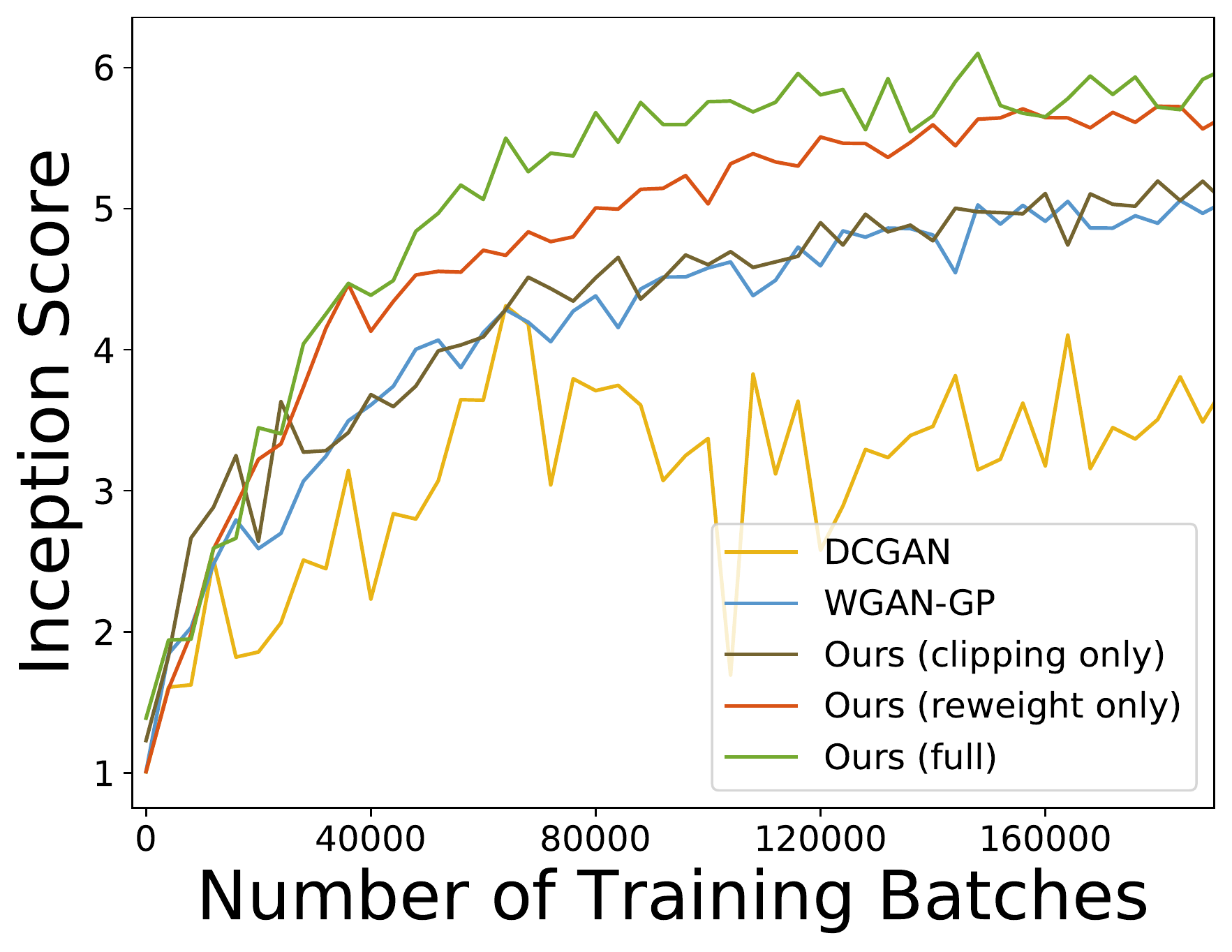}
\hspace{7pt}
\includegraphics[width=0.41\columnwidth]{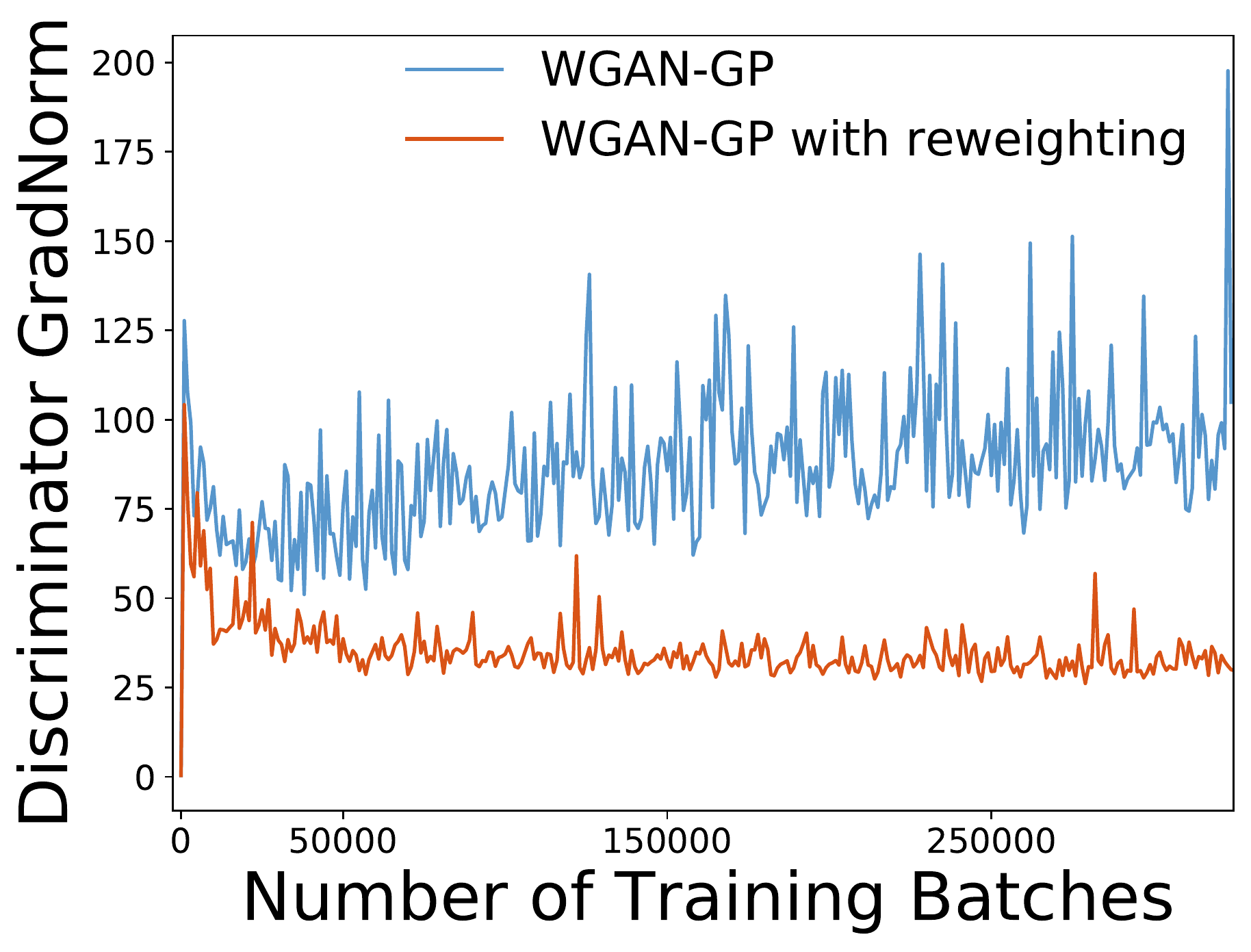}
%}
%\vskip -0.1in
\vspace{-2pt}
\caption{{\bf Left:}  Inception score on CIFAR-10 v.s. training batches (including both generator and discriminator batches). The DCGAN~\cite{DCGAN} architecture is used. {\bf Right:} The gradient norms of discriminators on fake samples. 
%The x-axis accounts for the training batches of both the generator and the discriminator.
%Sample re-weighting leads to much lower variance of the gradients.
}
\label{fig:image:converge}
\end{center}
%\vskip -0.4in
\vspace{-5pt}
\end{figure}

%\vspace{-1.2em}
\subsection{Image Generation}\label{sec:exp:image}
We first use the popular CIFAR-10 benchmark for evaluation and in-depth analysis of our approach.

%We also benchmark our model on unsupervised image generation with CIFAR10, our model outperforms the previous state-of-the-art by a large margin both qualitatively and quantitatively according to Inception Score and FID.

% \begin{figure}[ht]
% % \vskip 0.05in
% \begin{center}
% \centerline{\includegraphics[width=0.9\columnwidth]{cifar10/6.jpg}}
% \vskip -0.15in
% \caption{More samples generated by our generator on unsupervised image generation on CIFAR10}
% \label{samples}
% \end{center}
% \vskip -0.3in
% \end{figure}

\textbf{Setup.}  
%Here we evaluate the proposed method on image generation task to further demonstrate its effectiveness. This task requires the generator to transform random input vectors into images that resemble the images from the real dataset.
%CIFAR-10~\cite{cifar10} as our evaluation benchmark. This dataset consists of 50,000 natural images of size $32\times 32$ and is one of the most widely used dataset for image generation task. We implement a ResNet generator and discriminator according to the architecture guidelines offered by \cite{SNGAN}, and enforce Lipschitz constraint on the discriminator through gradient penalty \cite{wgangp} and consistency regularization \cite{wwgan}. 
CIFAR-10~\cite{cifar10} contains 50K images of sizes $32\times 32$. Following the setup in CT-GAN~\cite{wwgan}, 
%and its public implementation\footnote{\href{https://github.com/igul222/improved_wgan_training}{https://github.com/igul222/improved\_wgan\_training}}, 
we use a residual architecture to implement both generator and discriminator, and also impose a Lipschitz constraint on the discriminator. For each iteration, we update both generator and discriminator for 5 times. We use Inception Score (IS)~\cite{salimans2016improved} for evaluating generation quality and diversity, and Frechet Inception Distance (FID)~\cite{FID} for capturing model issues, e.g., mode collapse~\cite{xu2018empirical}. 

\textbf{Results.} Table~\ref{tab:exp:image:IS_FID} reports the results on CIFAR-10. 
For the three latest methods, SN-GANs~\cite{SNGAN} introduced spectral normalization to stabilize the discriminator training; WGAN-ALP~\cite{wganalp} developed an explicit Lipschitz penalty; and SRNGAN~\cite{srngan} introduced a weight-normalization scheme for generalization.
%and AutoGAN~\cite{gong2019autogan} incorporated neural architecture search for the generator architecture. 
Table~\ref{tab:exp:image:IS_FID} shows that our full approach (CT-GAN + discriminator sample re-weighting + generator probability ratio clipping) achieves the best, with both IS and FID significantly surpassing the  baselines. These results accord with the visual results in Figure~\ref{cifar10_combined} where our generated samples show higher visual quality than those of the baselines. Comparison between CT-GAN and our approach with only re-weighting shows significant improvement. By further adding the probability ratio clipping to arrive our full approach, the performance (both IS and FID) is further improved with a large margin. The results demonstrate the effectiveness of the two components in our approach.

% We also report the variant of our approach that applies only the discriminator re-weighting (sect~\ref{sec:method:dis}). Comparison with the base model CT-GAN~\cite{wwgan} shows that the re-weighting alone enables strong improvement in terms of Inception Score. By further applying clipped surrogate loss to the generative objective (sect~\ref{sec:method:gen}) and arriving the full approach, we achieve further improvement by a large margin.

%\hzt{Example images}

Figure~\ref{fig:intro-stability} in Sec.\ref{sec:intro} has shown the effects of the proposed approach in stabilizing the generator and discriminator training. Here we further analyze these two components. Figure~\ref{fig:image:converge} (left) shows the convergence curves of different GAN methods. For a fair comparison, all models use the same DCGAN architecture~\cite{DCGAN}, and both our approach and WGAN-GP~\cite{wgangp} enforce the same discriminator Lipschitz constraint. Following the optimal setup in \citep{wgangp}, the update ratio of both WGAN-GP and our ``re-weighting only'' is 5:1 (i.e., each iteration updates the discriminator for 5 times and the generator for one time). Our full approach and ``clipping only'' use an update ratio of 5:5, because the probability ratio clipping that discourages large generator updates allows us to update the generator more frequently, which is desirable. Note that the x-axis in Figure~\ref{fig:image:converge} accounts for both generator and discriminator batches (i.e., an 5:5 iteration is counted as 10 training batches). Thus, for any given point on the x-axis, all comparison methods used roughly the same amount of computation.
%%%
From the curves, one can observe that our full approach surpasses our approach with only sample re-weighting, and they both converge faster and achieve a higher IS score than ``clipping only'', WGAN-GP, and DCGAN. It is interesting to note that ``clipping only'' does not offer a performance improvement over WGAN-GP, though its combination with sample re-weighting (i.e., the full approach) does improve over ``re-weighting only''. This is indeed not unexpected, because clipping and re-weighting are derived from the variational framework (Eq.\ref{eq:var-obj}) in a principled way. Discarding either of the two could lead to improper handling of the variational distribution $q$ and fails to conform to the framework. 

Figure~\ref{fig:image:converge} (right) investigates how the fake sample re-weighting can affect the discriminator training. By injecting re-weighting into WGAN-GP, the gradients on fake samples become more stable with lower variance, which partially explains the better training stability of discriminator in Figure~\ref{fig:intro-stability}.

\begin{table}[t]
\renewcommand{\arraystretch}{0.9} 
\centering
\small
\begin{tabular}{l | llllll | l}
\toprule
Length & MLE  & SeqGAN~\cite{yu2017seqgan} & LeakGAN~\cite{guo2018long} & RelGAN~\cite{nie2018relgan}  & WGAN-GP~\citep{wgangp} & Ours & Real  \\
\cmidrule{1-8}
20   & 9.038 & 8.736   & 7.038   & 6.680 & 6.89  & \textbf{5.67} & 5.750  \\
40   & 10.411 & 10.310  & 7.191   & 6.765 & 6.78 & \textbf{6.14} & 4.071\\
\bottomrule
\end{tabular}
\vspace{1pt}
\caption{Oracle negative log-likelihood scores ($\downarrow$) on synthetic data. 
%Results of previous text GANs are from~\cite{nie2018relgan}.
} \label{tab:txtgan_oracle}
\vspace{-10pt}
\end{table}

% \begin{table}[H]
% \centering
% \small
% \begin{tabular}{l | lllllll | l}
% \toprule
% Length & MLE  & SeqGAN~\cite{yu2017seqgan} & RankGAN & LeakGAN & RelGAN  & WGANGP & Ours & Real  \\
% \cmidrule{1-9}
% 20     & 9.038 & 8.736  & 8.247   & 7.038   & 6.680 & 6.89  & \textbf{5.67} & 5.75  \\
% 40     & 10.411 & 10.31  & 9.958   & 7.191   & 6.765 & 6.78     & \textbf{6.14} & 4.071\\
% \bottomrule
% \end{tabular}
% \caption{Negative log likelihood our proposed model v.s. baselines on the oracle dataset.} \label{txtgan_oracle}
% \end{table}

\begin{table}[t]
\renewcommand{\arraystretch}{0.9} 
\setlength{\tabcolsep}{4.9pt} 
\small
\centering
\begin{tabular}{@{}r llll | l | l@{}}
\toprule
{\bf Method}        & {\bf BLEU-2} ($\uparrow$) & {\bf BLEU-3} ($\uparrow$) & {\bf BLEU-4} ($\uparrow$) & {\bf BLEU-5} ($\uparrow$) & {\bf NLL$_{gen}$} ($\downarrow$) & {\bf Human} ($\uparrow$)\\
\cmidrule{1-7}
MLE           & 0.768  & 0.473  & 0.240   & 0.126  & 2.382  & - \\
%SeqGAN~\cite{yu2017seqgan}        & 0.777  & 0.491  & 0.261  & 0.138  & 2.773  & - \\
%RankGAN       & 0.727  & 0.435  & 0.209  & 0.101  & 3.345  \\
LeakGAN~\cite{guo2018long}       & 0.826  & 0.645  & 0.437  & 0.272  & 2.356  & - \\
RelGAN 100~\cite{nie2018relgan}  & 0.881  & \textbf{0.705}  & \textbf{0.501}  & 0.319  & 2.482  & - \\
RelGAN 1000~\cite{nie2018relgan} & 0.837  & 0.654  & 0.435  & 0.265  & 2.285  & 3.42$\pm$1.23 \\
WGAN-GP~\cite{wgangp}       & 0.872  & 0.636  & 0.379  & 0.220   & \textbf{2.209}  & - \\\cmidrule{1-7}
Ours          & \textbf{0.905}  & 0.692  & 0.470   & \textbf{0.322}  & 2.265 & {\bf 3.59} $\pm$ {\bf 1.12} \\
\bottomrule
\end{tabular}
\vspace{1pt}
\caption{
Results on EMNLP2017 WMT News. BLEU measures text quality and NLL$_{gen}$ evaluates sample diversity. Results of previous text GAN models are from~\cite{nie2018relgan}, where RelGAN (100) and RelGAN (1000) use different hyper-parameter for gumbel-softmax. Our approach uses the same gumbel-softmax hyper-parameter as RelGAN~(1000). 
}
\label{tab:text-emnlp}
\vspace{-10pt}
\end{table}

\subsection{Text Generation}\label{sec:exp:text}
In this section, we evaluate our approach on text generation, a task that is known to be notoriously difficult for GANs due to the discrete and sequential nature of text. 

\textbf{Setup.} We implement our approach based on the RelGAN~\cite{nie2018relgan} architecture, a state-of-the-art GAN model for text generation. Specifically, 
we replace the generator and discriminator objectives in RelGAN with ours. We follow WGAN-GP~\cite{wgangp} and impose discriminator Lipschitz constraint with gradient penalty. 
Same as~\cite{nie2018relgan}, we use  Gumbel-softmax approximation~\cite{jang2016categorical,maddison2016concrete} on the discrete text to enable gradient backpropagation, and the generator is initialized with maximum likelihood (MLE) pre-training. 
%Our implementation is based on the public PyTorch code of RelGAN\footnote{\href{https://github.com/williamSYSU/TextGAN-PyTorch}{https://github.com/williamSYSU/TextGAN-PyTorch}}.
Same as previous studies, we evaluate on both synthetic and real text datasets.

\textbf{Results on Synthetic Data.} 
The synthetic data consists of 10K discrete sequences generated by an oracle-LSTM with fixed parameters~\cite{yu2017seqgan}. This setup facilitates evaluation, as the quality of generated samples can  be directly measured by  the negative log-likelihood (NLL) of the oracle on the samples. We use synthetic data with sequence lengths 20 and 40, respectively. Table~\ref{tab:txtgan_oracle} reports the results. MLE is the baseline with maximum likelihood training, whose output model is used to initialize the generators of GANs. Besides the previous text generation GANs~\cite{yu2017seqgan,guo2018long,nie2018relgan}, we also compare with WGAN-GP which uses the same neural architecture as RelGAN and ours. From Table~\ref{tab:txtgan_oracle}, one can observe that  our approach significantly outperforms all other approaches on both synthetic sets. Our improvement over RelGAN and WGAN-GP demonstrates that our proposed generator and discriminator objectives are more effective than the previous ones.

\textbf{Results on Real Data.}
We then evaluate our method on the EMNLP2017 WMT News, a large real text data used for text GAN studies~\cite{guo2018long,nie2018relgan}. The dataset consists of 270K/10K training/test sentences with a maximum length of 51 and a vocabulary size of 5,255. To measure the generation quality, we use the popular BLEU-$n$ metric which measures $n$-gram overlap between generated and real text ($n\in\{2,3,4,5\}$). To evaluate the diversity of generation, we use the negative log-likelihood of the generator on the real test set (NLL$_{gen}$)~\cite{guo2018long,nie2018relgan}. From the results in Table~\ref{tab:text-emnlp}, one can see that our approach shows comparable performance with the previous best model RelGAN~(100) in terms of text quality (BLEU), but has better sample diversity.
Our model also achieves much higher BLEU scores than WGAN-GP. 
%(e.g., 0.322 v.s. 0.220 on BLEU-5), demonstrating its ability of generating higher-quality samples. 
% \hzt{Human evaluation setup ...}
We perform \emph{human} evaluation, with randomly sampled 50 sentences for RelGAN~(1000) against ours and asked 5 annotators to score each sentence on a scale of 1-5. We use the same questions as designed by \citep{nie2018relgan}. Ours obtained an average human score of $3.59 \pm 1.12$, higher than $3.42 \pm 1.23$ by RelGAN (Fleiss' Kappa score $0.61$ showing substantial inter-rater agreement).

\begin{figure*}
\begin{minipage}{0.34\textwidth}
% \begin{table}[H]
%  \vskip - 0.23in
%\vspace{-1.0em}
%\setlength{\tabcolsep}{2.9pt}  
%\renewcommand{\arraystretch}{0.7} 
\small
  \centering
  \begin{tabular}{@{}rl@{}}
    \toprule
    {\bf Method} &  {\bf BLEU} \\
    \cmidrule{1-2}
    \citet{zhang2018style} & 24.48 \\
    \citet{tian2018structured} & 24.90 \\
    \citet{subramanian2018multiple} & 31.20 \\
    \citet{tikhonov2019style} & 32.82 \\\cmidrule{1-2}
    Ours & \textbf{33.45$\pm$.95}\\
    \bottomrule
  \end{tabular}
  \captionof{table}{
 BLEU scores between model generations and human-written text on the Yelp data. We run our method for 5 times and report the average and standard deviation.
%BLEU between output and human-written reformulations of Yelp! reviews, measured after five re-runs. Our method outperforms the baseline \cite{tikhonov2019style} as well as other state of the art models.
  } %TODO
  \label{tab:styletransfer_tbl2}
%   \vskip -0.2in
% \end{table}
\vspace{-20pt}
\end{minipage}
\hfill
\begin{minipage}{0.64\textwidth}
% \begin{figure}[t]
% \vskip 0.1in
\begin{center}
\centerline{\includegraphics[width=0.85\columnwidth]{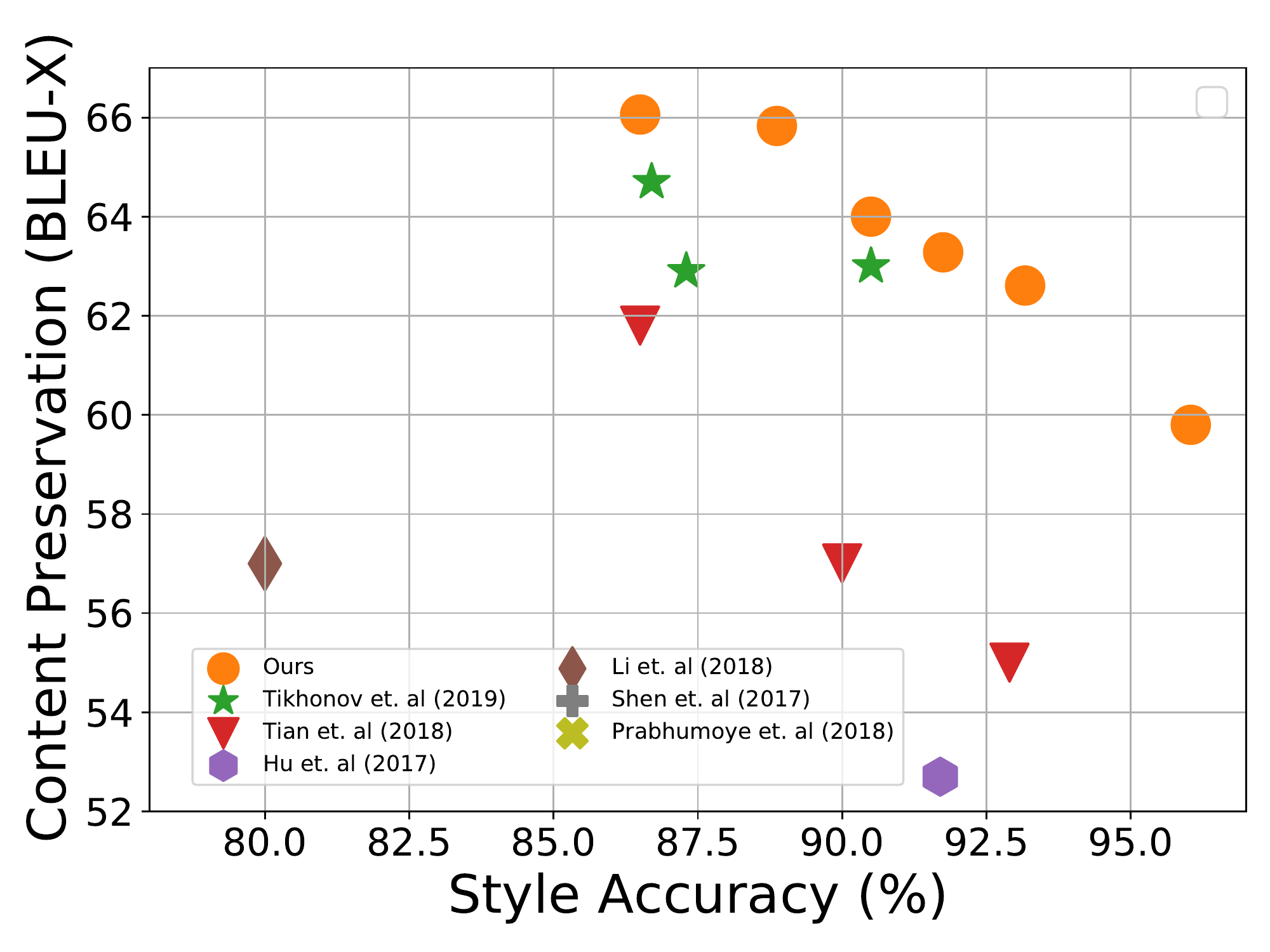}}
\vspace{-10pt}
\captionof{figure}{
Trade-off between style accuracy and content preservation. The {\color{orange} orange} circles denote our results using varying values for an objective weight~\cite{tikhonov2019style} which manages the trade-off. 
%Overview of the self-reported results for sentiment transfer on Yelp! reviews alongside with the results for the baseline model \cite{tikhonov2019style} averaged after five re-trains. The $6$ square points represent $6$ different weights for the classifier loss, which manages the trades-off between BLEU and accuracy. 
%
%Our results shown on the top-right of the graph, achieves higher overall accuracy-BLEU trade-off compared to the previous state-of-the-art methods.
%We did not report error margin the performance of our architecture is quite consistent. Results marked as $\nabla$ have BLEU score below $30$.
}
\label{fig:styletransfer_img}
\end{center}
%\vskip -0.4in
% \end{figure}
\end{minipage}
\vspace{-20pt}
\end{figure*}

\subsection{Text Style Transfer}\label{sec:exp:style}
Text style transfer task is gaining increasing attention in NLP~\cite{hu2017toward,shen2017style,yang2018unsupervised}. The task aims at rewriting a sentence to modify its style (e.g., sentiment) while preserving the content. Previous work applies GANs on neural hidden states to learn disentangled representations~\cite{shen2017style,tikhonov2019style}. The task thus can serve as a good benchmark for GANs, as hidden state modeling provides a new modality that differs from image and text modeling as studied above.

%We are able to improve upon the state-of-the-art text style transfer model simply by adapting our GAN formulation to the style discriminator.
% apply our framework to a unsupervised text style transfer model described in, using the Texar-Pytorch \cite{texar} framework.
% \begin{figure}[ht]
% \vskip 0.2in
% \begin{center}
% \centerline{\includegraphics[width=0.5\columnwidth]{style_transfer/Style_transfer.png}}
% \vskip -0.2in
% \caption{The generative model, where the style discriminator ($D$) is a structured constraint the generator optimize against. A latent code discriminator ensure the independence between semantic part of the latent representation and the style of the text. Blue dashed arrows denote additional independence constraints of latent representation and controlled attribute, see \cite{tikhonov2019style} for the details.}
% \label{styletransfer_model}
% \end{center}
% % \vskip -0.3in
% \end{figure}

\textbf{Setup.} We follow the same experimental setting and use the same model architecture in the latest work~\cite{tikhonov2019style}. In particular, the VAE-based model~\cite{hu2017toward,kingma2013auto} is extended by adding a latent code discriminator which eliminates stylistic information in the latent code. We replace their adversarial objectives with our proposed ones, and impose discriminator Lipschitz constraint with gradient penalty~\cite{wgangp}. 
%Our implementation is based on the public code\footnote{\href{https://github.com/VAShibaev/text_style_transfer}{https://github.com/VAShibaev/text\_style\_transfer}} released in \cite{tikhonov2019style}. 
We test on sentiment transfer, in which the sentiment (positive/negative) is treated as the text style. We use the standard Yelp review dataset,
%\footnote{\href{https://www.yelp.com/dataset}{www.yelp.com/dataset}}, 
and the ground truth output text provided by \cite{li2018delete}.
%Same as prior works, we conduct our experiments on the Yelp! reviews dataset\footnote{\href{https://www.yelp.com/dataset}{www.yelp.com/dataset}}, which consists of Yelp! reviews labeled as negative or positive, enhanced with human written reference sentiment reformulations by \cite{li2018delete}.

\textbf{Results.} Following the previous work~\cite{tikhonov2019style}, we first report the BLEU score that measures the similarity of the generated samples against the human written text. Table~\ref{tab:styletransfer_tbl2} shows that our approach achieves best performance, improving the state-of-the-art result~\cite{tikhonov2019style} from BLEU $32.82$ to $33.45$.

The second widely used evaluation method is to measure (1) the style accuracy by applying a pre-trained style classifier on generated text, and (2) the content preservation by computing the BLEU score between the generated text and the original input text (BLEU-X). There is often a trade-off between the two metrics. Figure~\ref{fig:styletransfer_img} displays the trade-off by different models. Our results locate on the top-right corner, indicating that our approach achieves the best overall style-content trade-off.

\section{Conclusion}\label{sec:conclusion}
We have presented a new training framework of GANs derived from a new variational perspective and draws on rich connections with RL-as-inference. This results in probably ratio clipping for generator updates to discourage overly large changes, and fake sample re-weighting for stabilized discriminator updates. Experiments show our approach demonstrates superior training stability and improves over previous best methods on image generation, text generation, and text style transfer. 
The connection between the GAN and RL formalisms can potentially inspire more cross-pollination between the two fertile research fields. We are also interested in extending the formulation to connect more machine learning paradigms~\citep{hu2020learning}, for more systematic understanding, unification, and generalization of diverse learning algorithms.
%We are interested in exploring more connections between GANs and other learning paradigms to inspire more techniques for improved GAN training.

% This paper proposes a new perspective of GAN training from posterior regularization, which results in two novel revisions to the previous GANs formulations: 1) in the phase of discriminator learning, a more effective sampling mechanism for re-weighting fake samples, and 2) in the phase of generator learning, a relative entropy regularization for stabilizing the updates. We prove that the new formulation is compatible with the latest theoretical guarantees on informative gradients and convergence. Experiments show that our framework obtains the state-of-the-art FID and Inception score on CIFAR-10 dataset for unsupervised image generation, and state-of-the-art overall performance in text-style-transfer according to BLEU score and style discriminator accuracy.

%\newpage

\section*{Broader Impacts}
This work offers a unique viewpoint on two promising fields with lots of applications and impacts: Generative Adversarial Networks and Reinforcement Learning. The improvement to image generation results may be adapted to speed up photo editing, improve scene rendering, and create more realistic simulation for robot training. Furthermore, the contribution to text generation and text style transfer can be adopted to improve the quality of machine translation, and automated news-summaries.

Nevertheless, GANs can also be applied to faking images of people and jeopardize personal identities (i.e. Deepfake). We hope that future works can counter this issue through deep-fake detection.
\bibliography{citations}

% \bibliographystyle{ieee}
% \section*{References}

% References follow the acknowledgments. Use unnumbered first-level heading for
% the references. Any choice of citation style is acceptable as long as you are
% consistent. It is permissible to reduce the font size to \verb+small+ (9 point)
% when listing the references.
% {\bf Note that the Reference section does not count towards the eight pages of content that are allowed.}
% \medskip

% \small

% [1] Alexander, J.A.\ \& Mozer, M.C.\ (1995) Template-based algorithms for
% connectionist rule extraction. In G.\ Tesauro, D.S.\ Touretzky and T.K.\ Leen
% (eds.), {\it Advances in Neural Information Processing Systems 7},
% pp.\ 609--616. Cambridge, MA: MIT Press.

% [2] Bower, J.M.\ \& Beeman, D.\ (1995) {\it The Book of GENESIS: Exploring
%   Realistic Neural Models with the GEneral NEural SImulation System.}  New York:
% TELOS/Springer--Verlag.

% [3] Hasselmo, M.E., Schnell, E.\ \& Barkai, E.\ (1995) Dynamics of learning and
% recall at excitatory recurrent synapses and cholinergic modulation in rat
% hippocampal region CA3. {\it Journal of Neuroscience} {\bf 15}(7):5249-5262.

\newpage

\input{appendix}

\end{document}

%% file: MACPdef_manu.tex
\newcommand{\led}[1]{\overset{\text{\ding{#1}}}{\leq}}

\newcommand{\lee}[1]{\overset{\text{\ding{#1}}}{=}}

\newcommand{\kl}{\textbf{KL}}

\newcommand{\wm}{\bm{\theta}}
\newcommand{\xm}{\bm{x}}
\newcommand{\ym}{\bm{y}}

\newcommand{\EE}{\mathbb{E}}
\newcommand{\PP}{\mathbb{P}}

\newcommand{\norm}[1]{\left\|#1\right\|}

%% file: appendix.tex
\section{Appendix}

\subsection{Proof on the equivalence between Reverse KL Divergence and KL Divergence} \label{pf_kl}
We prove that optimizing $\kl(p_{\wm} || q )$ are equivalent to optimizing $\kl(q || p_{\wm})$.
%when the function $f_{\phi}(\xm)$ is lower and upper-bounded. 
This provides guarantee for the approximation that leads to \eqref{eq:g-loss}. % in~\cite{hu2018deep}.

\textbf{Claim:} Under the assumption that $f_\phi$ Lipschitz, $f_\phi$ is bounded because the input $\xm$ is bounded. Let $K$ be the Lipschitz constant of $f_\phi$, and let $c = f_\phi(0)$
\begin{equation}
    |f_\phi(x)-c|\leq K|x-0| = K|x|
\end{equation}
We then show that $\kl(p_{\wm} || q )$ differ $\kl(q || p_{\wm})$ by at most a constant. Since the function $f_{\phi}(\xm)$ is lower and upper-bounded. There exists $a,b$, such that $-a \leq f_{\phi}(\xm) \leq b$ for any $\xm$ bounded.
\begin{equation}\label{asfsafvsgvsdf}
\begin{split}
 & \kl(q || p_{\wm}) - \kl(p_{\wm} || q )  \\
= & \int_{\xm} \left[ q(\xm) \log\left(\frac{q(\xm)}{p_{\wm}(\xm)}\right)- p_{\wm}(\xm) \log\left(\frac{p_{\wm}(\xm)}{q(\xm)}\right) \right] d \xm\\
= & \int_{\xm} \left[ q(\xm) +p_{\wm}(\xm) \right] \log\left(\frac{q(\xm)}{p_{\wm}(\xm)}\right)  d \xm\\
\lee{172} & \int_{\xm}p_{\wm}(\xm)  \left[ 1 +  \frac{\exp{(\alpha f_{\phi}(\xm))}}{Z} \right] \log\left( \frac{\exp{(\alpha f_{\phi}(\xm))}}{Z}\right)  d \xm\\
\led{173} & \alpha(a+b) \int_{\xm}p_{\wm}(\xm)  \left[ 1 +  \frac{\exp{(\alpha f_{\phi}(\xm))}}{Z} \right]    d \xm\\
\lee{174} & 2\alpha(a+b),\\
\end{split}
\end{equation}
where \ding{172} plugs $q^*(\xm) = \frac{p_{\wm}(\xm)\exp{(\alpha f_{\phi}(\xm))}}{Z}$; \ding{173} uses the fact $\log\left( \frac{\exp{(\alpha f_{\phi}(\xm))}}{Z}\right) = \log\left( \frac{\exp{(\alpha f_{\phi}(\xm))}}{\int_{\xm} p_{\wm}(\xm)\exp{(\alpha f_{\phi}(\xm))} d \xm }\right) \leq \log\left( \frac{\exp{(\alpha b)}}{\int_{\xm} p_{\wm}(\xm)\exp{(-\alpha a)} d \xm }\right)= \alpha(a+b)$; \ding{174} uses $\int_{\xm}p_{\wm}(\xm)     \frac{\exp{(\alpha f_{\phi}(\xm))}}{Z} d \xm=1$.
% Note that in~\eqref{qsolution}, for optimal $f^*_{\phi}(\xm)$ we have
% \begin{equation}\label{afasdcsadc}
% \begin{split}
% \frac{p_{\wm}(\xm)}{q(\xm)} =  \frac{Z}{\exp{(\alpha f^*_{\phi}(\xm))}}.
% \end{split}
% \end{equation}
% So $\frac{p_{\wm}(\xm)}{q(\xm)}$ has a constant value which straightforwardly implies $p_{\wm} =q$. This is because optimal $f^*_{\phi}(\xm)$ is a constant for any $\xm$.
% @Pan. Thanks!
The above claim completes the theoretical guarantee on the reverse-KL approximation in \eqref{eq:g-loss}.

\subsection{Proof on the necessity of Lipschitz constraint on the discriminator} \label{pf_issue}
Although \cite{hu2018deep} shows preliminary connections between PR and GAN, the proposed PR framework does not provide informative gradient to the generator when treated as a GAN loss. 
Following~\cite{zhou2018understanding}, %we provide certain explanation for the proposed method.
% \cite{zhou2018understanding} focuses on the 
we consider the training problem when the discriminator (i.e. $f_{\phi}(\xm)$ here) is optimal: when discriminator $f^*_{\phi}(\xm)$ is optimal, then the gradient of generator $g(f_{\phi}(\xm))$ is $\nabla_{f^*_{\phi}(\xm)} g(f^*_{\phi}(\xm)) \cdot \nabla_{\xm} f^*_{\phi}(\xm)$ which could be very small due to vanished $\nabla_{\xm} f^*_{\phi}(\xm)$. In this way, it is hard to push the generated data distribution $p_{\wm}$ towards the targeted real distribution $p_r$. This problem also exists in \eqref{eq:d-grad} because
\begin{equation}%\label{afasdcsadc}
\begin{split}
f^*_{\phi}(\xm) =  \arg\min_{f_{\phi}(\xm)} \alpha \left[   p_r(\xm) f_{\phi}(\xm)  - q(\xm)    f_{\phi}(\xm) \right].
\end{split}
\end{equation}
So if $p_r$ and $q$ are disjoint, we have
\begin{equation}%\label{afasdcsadc}
\begin{split}
f^*_{\phi}(\xm)& =  \arg\min_{f_{\phi}(\xm)} \alpha \left[   p_r(\xm) f_{\phi}(\xm)  - q(\xm)    f_{\phi}(\xm) \right]\\
&= \begin{cases}
                \arg\min_{f_{\phi}(\xm)} p_r(\xm) f_{\phi}(\xm), & \mbox{if } \xm\sim p_r \\
                \arg\min_{f_{\phi}(\xm)} - q(\xm)    f_{\phi}(\xm), & \mbox{if } \xm \sim q.
                 \end{cases}
\end{split}
\end{equation}
Note that for any $\xm\sim p_r$, $f^*_{\phi}(\xm)$ is not related to $q$ and thus its gradient $\nabla f^*_{\phi}(\xm)$ also does not relate to $q$. Similarly, for any $\xm\sim q$, $\nabla f^*_{\phi}(\xm)$ does not provide any information of  $p_r$. Therefore, the proposed loss in \cite{hu2018deep} cannot guarantee informative gradient \cite{zhou2018understanding} that pushes $q$ or $p_{\wm}$ towards to $p_r$. 

\subsection{Experiments: More Details and Results}

\subsubsection{Binary classifier for probability ratio clipping}
For the image generation and text generation, the binary classifier $C$ in Eq.\eqref{eq:r-approx} has the same architecture as the discriminator except an additional Sigmoid activation at the output layer. The binary classifier is trained with real and fake mini-batches alongside the generator, and requires no additional loops. We select the clipping parameter $\epsilon$ from $\{0.2, 0.4\}$, as they are typically used in PPO.

In addition in the task of image generation, we observe similar overall performance between training on raw inputs from the generator/dataset and training on input features from the first residual block of the discriminator ($D$), thus further reducing the computational overhead of the binary classifier.

\subsubsection{Image Generation on CIFAR-10}
We translate the code\footnote{\href{https://github.com/biuyq/CT-GAN}{github.com/biuyq/CT-GAN}} provided by \citet{wwgan} into Pytorch to conduct our experiments. We use the same architecture: a residual architecture for both generator and discriminator, and enforcing Lipschitz constraint on the discriminator in the same way as CT-GAN~\cite{wwgan}. 
During training, we interleave 5 generator iterations with 5 discriminator iterations. We optimize the generator and discriminators with Adam (Generator lr: $5e-5$, Discriminator lr: $1e-4$, betas: $(0.0, 0.9)$). We set the clipping threshold $\epsilon := 0.4$ for the surrogate loss and we linearly anneal the learning rate with respect to the number of training epochs.
% The binary classifier $C$ \ref{eq:r-approx} has the same architecture as the discriminator. We observe similar overall performance training $C$ on raw inputs from the generator/dataset or on input features from the first residual block of the discriminator ($D$).

\paragraph{Discriminator sample re-weighting stabilizes DCGAN}
We quantitatively evaluate the effect of discriminator re-weighted sampling by comparing DCGAN~\cite{DCGAN} against DCGAN with discriminator re-weighting. Starting from the DCGAN architecture and hyper-parameters, we run 200 random configurations of learning rate, batch size, non-linearity (ReLU/LeakyReLU), and base filter count (32, 64). Results are summarized in Table \ref{tab:wwgan_tbl}. DCGANs trained with re-weighted sampling has significantly less collapse rate, and achieves better overall performance in terms of Inception Score. These results well demonstrate the effectiveness of the proposed discriminator re-weighted sampling mechanism.

\begin{table}[h]
  \vskip - 0.1in
  \centering
  \begin{tabular}{r lll}
    \toprule
    {\bf Method}    &  {\bf Collapse rate} & {\bf Avg IS} & {\bf Best IS}  \\
    \midrule
    DCGAN & 52.4\% & 4.2 & 6.1\\
    DCGAN + Re-weighting & 30.2\% &5.1 & 6.7\\
    \bottomrule
  \end{tabular}
\caption{Outcomes of 200 trials with random configurations. The performance of the models are measured through Inception score. We identify training collapse when the average discriminator loss over 2000 batches is below $1e^{-20}$ or above $1 - 1e^{-20}$. DCGAN re-weighted with our loss has lower collapse rate and higher average performance.}
  \label{tab:wwgan_tbl}
\end{table}

% \begin{figure}[h]
% % \vskip 0.1in
% \begin{center}
% \centerline{\includegraphics[width=\columnwidth]{cifar10/DCGAN_perf_small.jpg}}
% \vskip -0.22in
% \caption{DCGAN trained with or without discriminator weighted sampling. Two trials with different architectures are plotted. Generator loss marked as $\Delta$ when the loss explodes or vanishes due to the vanish and explosion of the discriminator loss.}
% \label{sample_from_q}
% \end{center}
% \vskip -0.38in
% \end{figure}
% To further illustrate how discriminator re-weighted sampling improves the training of an unstable GAN framework, we plot the generator losses of two different instance of DCGAN with and without discriminator re-weighted sampling under small batchsize (32, 64 respectively). The generator loss suffers from significantly more fluctuations in the original DCGAN formulation under small batchsize, but the re-weighted DCGANs still exhibit stable training curves.  
%All these results show that our discriminator re-weighted loss improves training stability, especially in scenarios where GAN suffers from high instability.

\begin{figure}[h]
% \vskip 0.2in
\begin{center}
\centerline{\includegraphics[width=0.7\columnwidth]{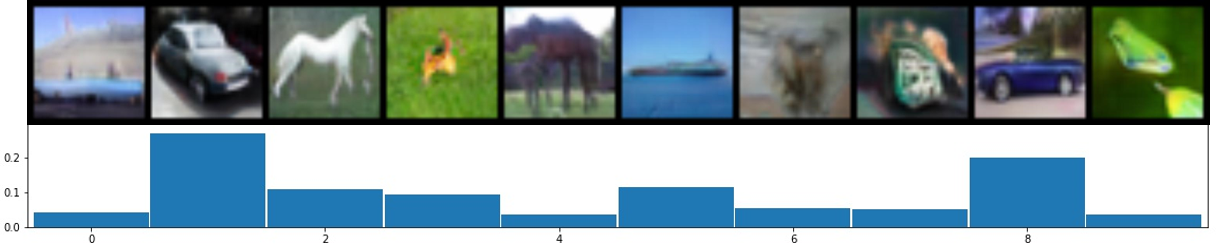}}
\vskip -0.1in
\caption{One batch of generated images together with their corresponding softmax discriminator weights. The more photo-realistic images (columns 2, 3, 5, 8) receive higher discriminator weights. In this batch, the generator will be influenced more by gradients from the better-quality samples above.}
\label{fig:q_weights}
\end{center}
\vskip -0.3in
\end{figure}

\paragraph{Discriminator re-weighted samples}  
To provide an illustration of how discriminator weights can help the discriminator concentrate on the fake samples of better quality during the training phase, in Figure~\ref{fig:q_weights} we plot the fake samples of a trained ResNet model alongside their corresponding discriminator weights. 

\paragraph{Clipped surrogate objective}
One unique benefit of the clipped surrogate objective is that it allows our model to obtain an estimate of the effectiveness of the discriminator, which then enables us to follow a curriculum that takes more than one $(n_g)$ generator steps per $(n_c)$ critic steps. In practice, setting $n_g = n_c = 5$ achieves good quality, which also allows us to take $5$ times more generator steps than prior works \cite{WGAN, wgangp, wwgan, SNGAN} with the same number of discriminator iterations.
Table~\ref{tab:exp:image:IS_FID} shows the improvement enabled by applying the surrogate objective.

\begin{figure}[t]
\vskip 0.2in
\begin{center}
\centerline{\includegraphics[width=.5\columnwidth]{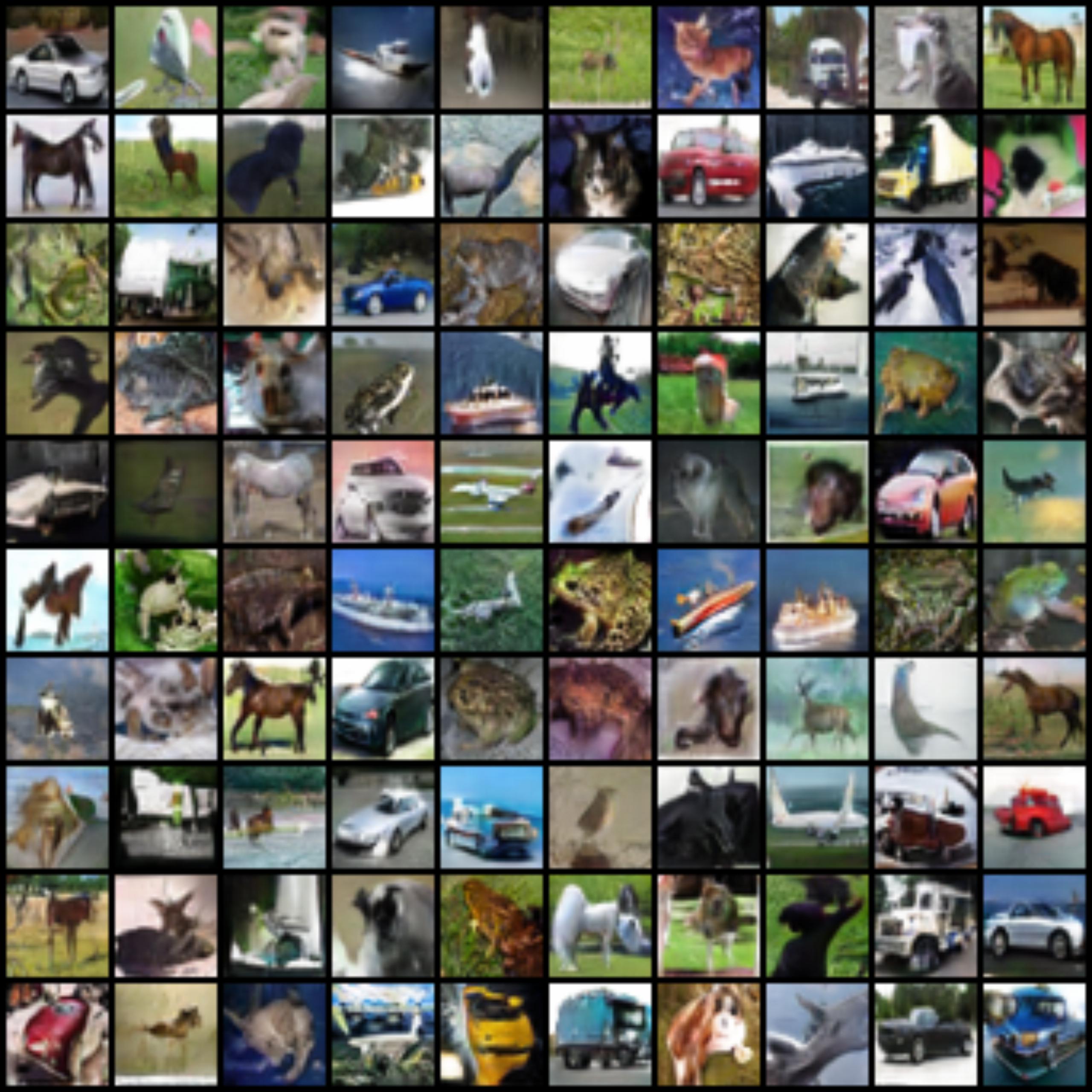}
            \includegraphics[width=.5\columnwidth]{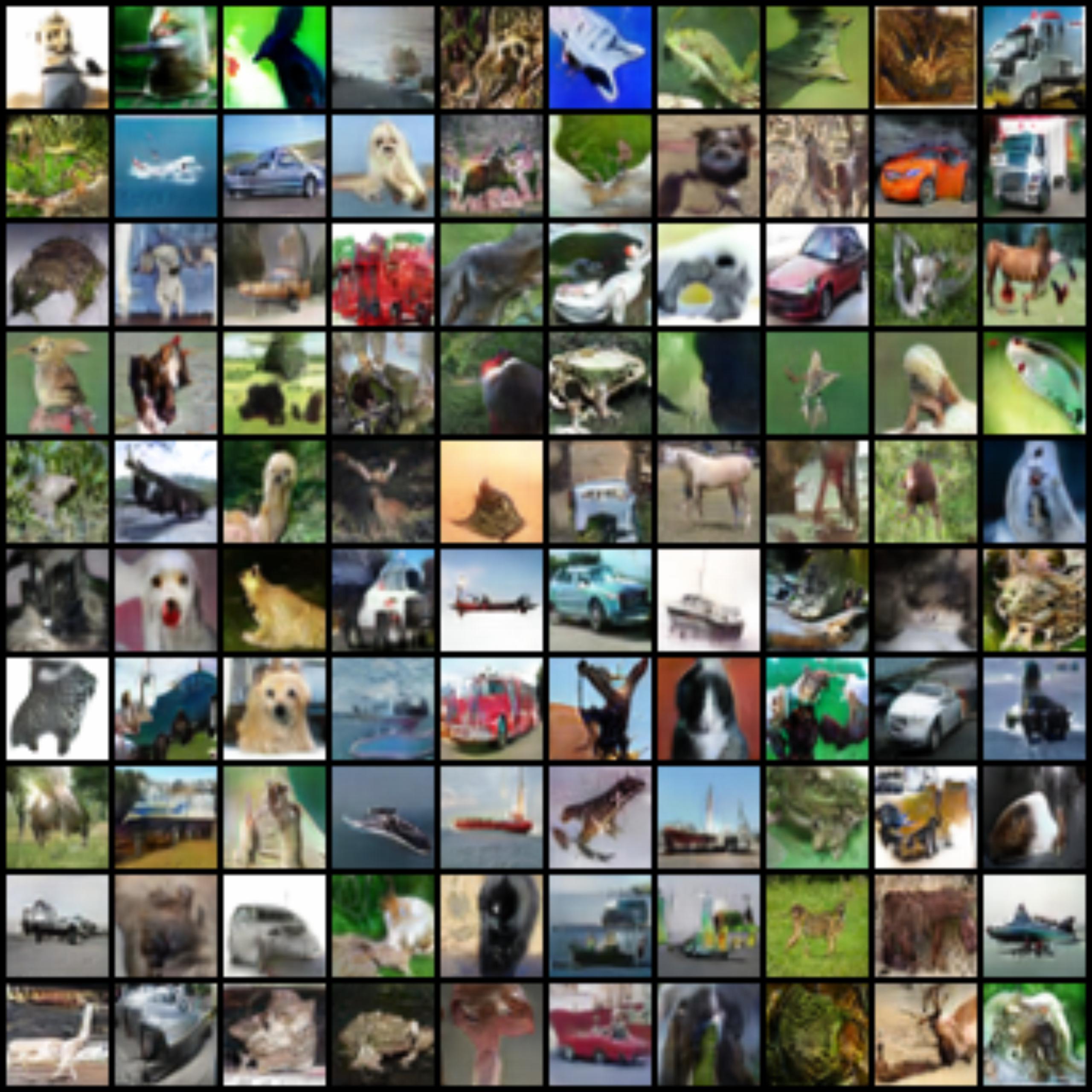}}
% \caption{More samples from our generator on CIFAR10 (unsupervised)}
\centerline{\includegraphics[width=.5\columnwidth]{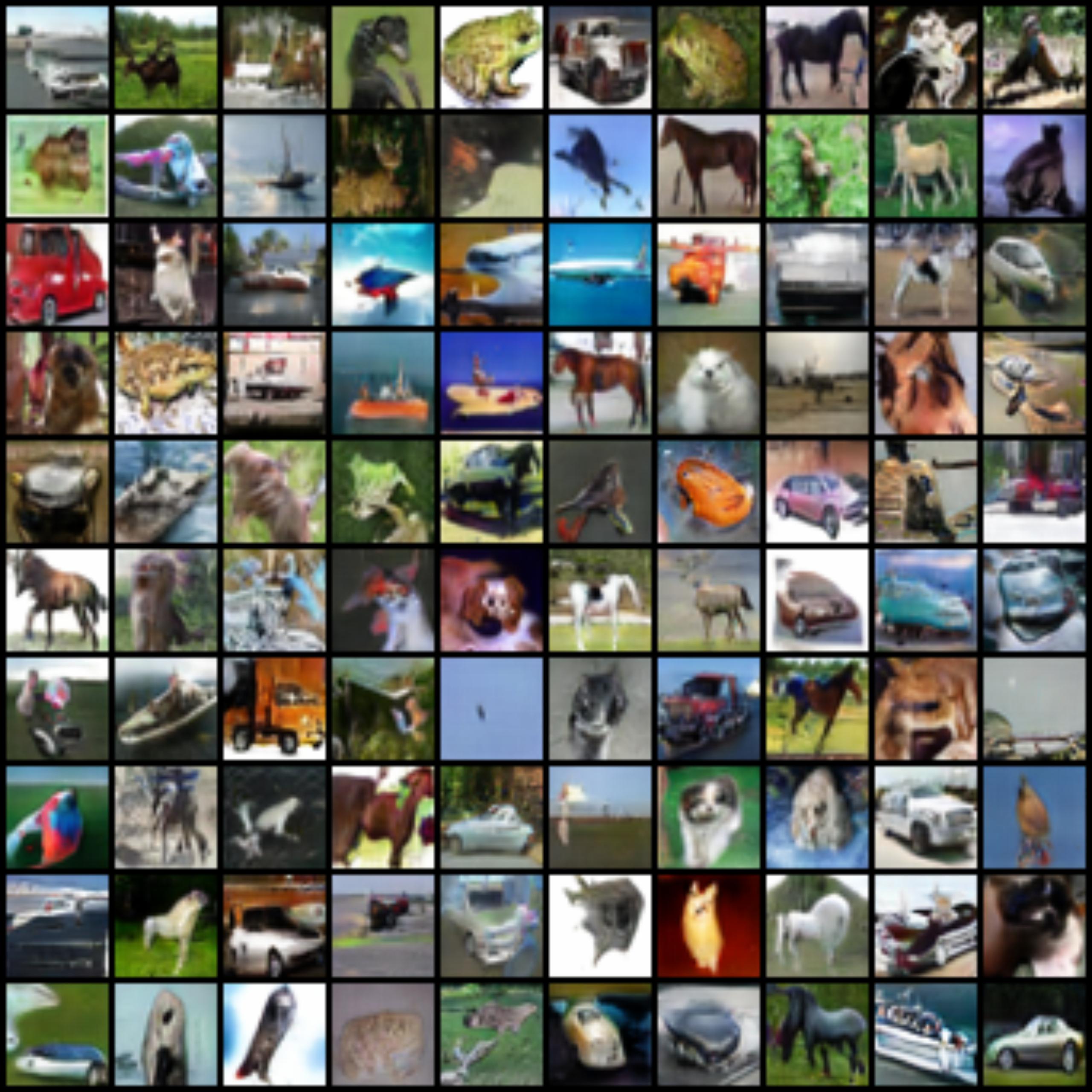}
            \includegraphics[width=.5\columnwidth]{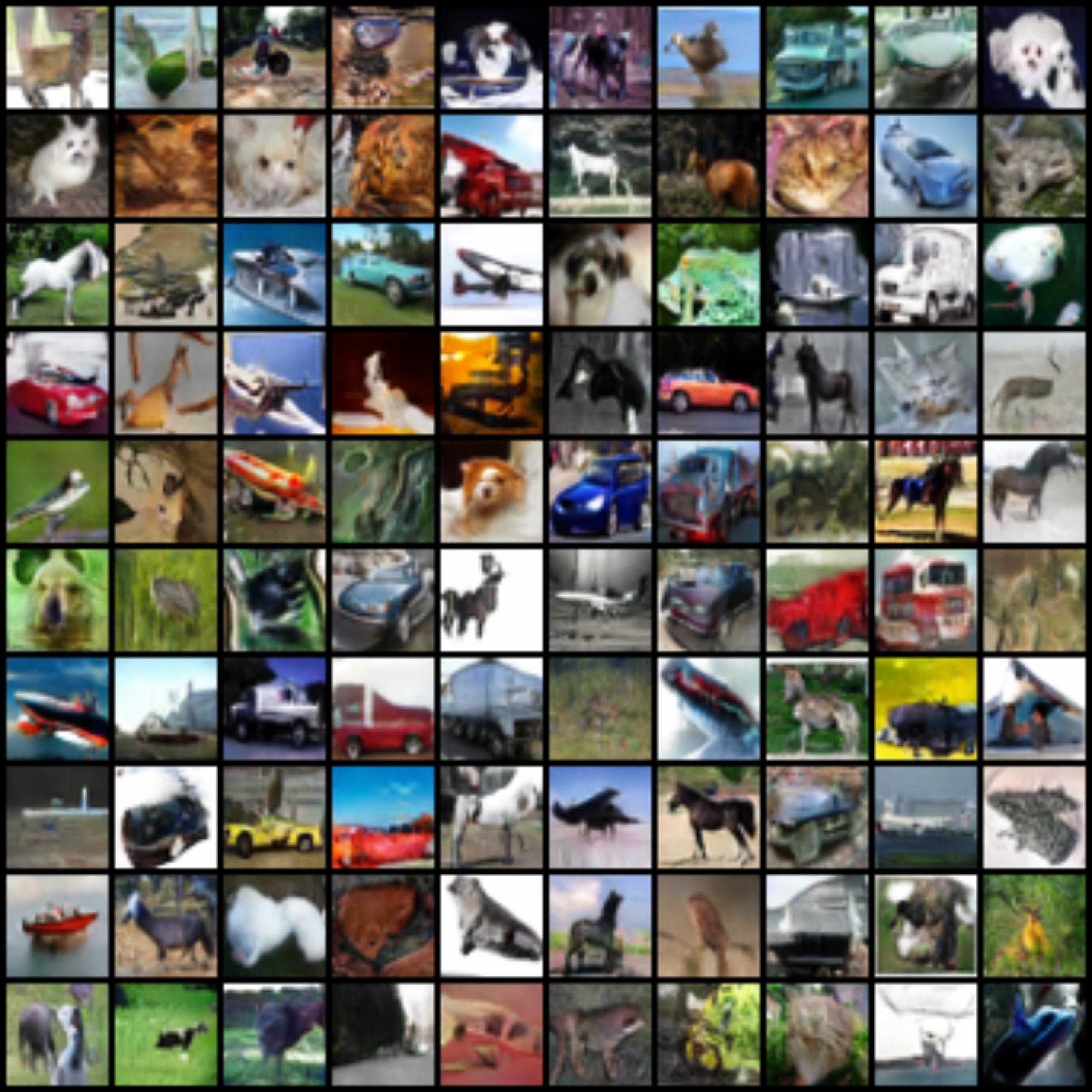}}
\caption{More samples from our generator on CIFAR-10}
\label{appendix:fig:samples}
\end{center}
\vskip -0.2in
\end{figure}

\paragraph{Generated samples} 
Figure~\ref{appendix:fig:samples} shows more image samples by our model.

\subsubsection{Text Generation}
We build upon the Pytorch implementation\footnote{\href{https://github.com/williamSYSU/TextGAN-PyTorch}{github.com/williamSYSU/TextGAN-PyTorch}} of RelGAN. We use the exact same model architecture as provided in the code, and enforce Lipschitz constraint on the discriminator in the same way as in WGAN-GP~\cite{WGAN}. 

During training, we interleave 5 generator iterations with 5 discriminator iterations. We use Adam optimizer (generator lr: 1e-4, discriminator lr: 3e-4). We set the clipping threshold $\epsilon=0.2$ for the surrogate loss and we linearly anneal the learning rate with respect to the number of training epochs.

\subsubsection{Text Style Transfer}
\begin{figure}[t]
\vskip 0.2in
\begin{center}
\centerline{\includegraphics[width=.5\columnwidth]{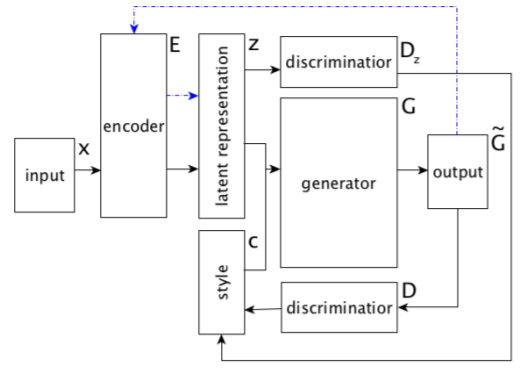}}
\caption{Model architecture from \cite{tikhonov2019style}, where the style discriminator ($D$) is a structured constraint the generator optimize against. A latent code discriminator ensure the independence between semantic part of the latent representation and the style of the text. \textcolor{blue}{Blue} dashed arrows denote additional independence constraints of latent representation and controlled attribute, see \cite{tikhonov2019style} for the details.}
\label{appendix:fig:style_transfer_arch}
\end{center}
\vskip -0.2in
\end{figure}

We build upon the Texar-TensorFlow~\cite{texar} style-transfer model by \citet{tikhonov2019style}\footnote{\href{https://github.com/VAShibaev/text_style_transfer}{https://github.com/VAShibaev/text\_style\_transfer}}. We use the exact same model architecture and hyper-parameters as provided in the code, and enforce Lipschitz constraint on the discriminator in the same way as WGAN-GP~\cite{WGAN}. In addition, we replace the discriminator $D$ in Figure \ref{appendix:fig:style_transfer_arch}, by our loss with an auxiliary linear style classifier as in \citet{ACGAN}. We did not apply the surrogate loss to approximate the KL divergence, but relied on gradient clipping on the generator.

% \paragraph{Results on COCO Image Captions dataset}
% Table~\ref{appendix:tab:text:coco} shows the results.
% %RelGAN equipped with our loss improves BLEU score (sample quality) by a large margin on the COCO Image Captions dataset, while maintaining comparable $NLL_{gen}$ score (sample diversity). 
% \begin{table}[H]
% \centering
% \small
% \begin{tabular}{r lllll}
% \toprule
% Method        & BLEU2        & BLEU3        & BLEU4        & BLEU5        &  $NLL_{gen}$     \\
% \midrule
% MLE           & 0.731         & 0.497         & 0.305         & 0.189         & 0.718         \\
% SeqGAN        & 0.745         & 0.498         & 0.294         & 0.18          & 1.082         \\
% RankGAN       & 0.743         & 0.467         & 0.264         & 0.156         & 1.344         \\
% LeakGAN       & 0.746         & 0.528         & 0.355         & 0.23          & 0.679         \\
% RelGAN (100)  & 0.849 & 0.687 & 0.502& 0.331 & 0.756\\
% RelGAN (1000) & 0.814 & 0.634 & 0.455 & 0.303 & \textbf{0.655} \\
% Ours          & \textbf{0.878 }        & \textbf{0.777}         & \textbf{0.549}         & \textbf{0.401}         & 1.170 \\       
% \bottomrule
% \end{tabular}
% \vspace{3pt}
% \caption{Results on the COCO Image Captions dataset.}
% \label{appendix:tab:text:coco}
% \end{table}